\documentclass[twoside,11pt]{article}

\usepackage{draftwatermark}
\SetWatermarkLightness{0.85}
\SetWatermarkScale{7}
\SetWatermarkText{arXiv}

\usepackage{multirow}
\usepackage{arxiv}
\usepackage{subfig}
\usepackage{url}
\usepackage{amsmath}
\usepackage{graphicx} 
\usepackage{bm}

\usepackage{arydshln}

\renewcommand{\bm}[1]{#1} 

\newcommand{\f}[1]{f_{\bm{\theta}}^{#1}}
\newcommand{\vidx}[2]{\left[ #1 \right]_{#2}}
\def\T{\mathrm{\scriptscriptstyle\bm{T}}}
\def\N{\mathbb{N}}
\def\R{\mathbb{R}}
\def\plusequal{\mbox{+=}}
\newcommand{\seq}[3]{[#1]_{#2}^{#3}}
\newcommand{\midx}[3]{\left[ #1 \right]_{#2,\,#3}}
\newcommand{\window}[3]{\langle#1\rangle_{#2}^{#3}}
\newcommand{\voireq}[1]{(\ref{#1})}
\newcommand{\voirsec}[1]{Section~\ref{#1}}
\newcommand{\voirfig}[1]{Figure~\ref{#1}}
\newcommand{\voirtbl}[1]{Table~\ref{#1}}
\newcommand{\col}[2]{\window{#1}{#2}{1}}

\def\eg.{\mbox{e.}\mbox{g.}}
\def\ie.{\mbox{i.}\mbox{e.}}
\DeclareMathOperator*{\logadd}{logadd}

\DeclareMathOperator*{\argmax}{argmax}

\jmlrheading{1}{2009}{1-48}{4/00}{10/00}{Ronan Collobert, Jason Weston, L\'eon Bottou, Michael Karlen, Koray Kavukcuoglu and Pavel Kuksa}

\ShortHeadings{Natural Language Processing (almost) from Scratch}{Collobert, Weston, Bottou, Karlen, Kavukcuoglu and Kuksa}
\firstpageno{1}

\begin{document} 

\title{Natural Language Processing (almost) from Scratch}

\author{\relax
 \name Ronan Collobert \email ronan@collobert.com\\
 \addr NEC Labs America, Princeton NJ. \\
 \name Jason Weston \email jweston@google.com\\
 \addr Google, New York, NY. \\
 \name L\'{e}on Bottou \email leon@bottou.org \\
 \name Michael Karlen \email michael.karlen@gmail.com\\
 \name Koray Kavukcuoglu$^{\dagger}$ \email koray@cs.nyu.edu \\
 \name Pavel Kuksa$^{\ddagger}$ \email pkuksa@cs.rutgers.edu \\
 \addr NEC Labs America, Princeton NJ.\\
}

\editor{}

\maketitle
{\def\thefootnote{$\dagger$}\footnotetext{Koray Kavukcuoglu is also with 
  New York University, New York, NY.}}
{\def\thefootnote{$\ddagger$}\footnotetext{Pavel Kuksa is also with 
  Rutgers University, New Brunswick, NJ.}}
\setcounter{footnote}{0}        

\raggedbottom
\sloppy
\hyphenpenalty=1000

\begin{abstract}%
We propose a unified neural network architecture and learning algorithm
that can be applied to various natural language processing tasks
including: part-of-speech tagging, chunking, named entity recognition,
and semantic role labeling.
This versatility is achieved by
trying to avoid task-specific engineering and therefore 
disregarding a lot of prior knowledge.
Instead of exploiting man-made
input features carefully optimized for each task, our system
learns internal representations on the basis of vast amounts
of mostly unlabeled training data.
This work is then used as a basis for building a freely available
tagging system with good performance and
minimal computational requirements.
\par\medskip
\end{abstract}



\section{Introduction}

Will a computer program ever be 
able to convert a piece of English text into a data structure 
that unambiguously and completely describes the meaning of the 
natural language text?
Among numerous problems, no consensus has emerged 
about the form of such a data structure.
Until such fundamental Artificial Intelligence problems are resolved, 
computer scientists must settle for reduced objectives:
extracting simpler representations describing 
restricted aspects of the textual information.  

These simpler representations are often motivated 
by specific applications, for instance, 
bag-of-words variants for information retrieval.  
These representations can also be motivated by our 
belief that they capture something more general 
about natural language. They can describe
syntactic information (\eg. part-of-speech tagging, 
chunking, and parsing) or semantic information
(\eg. word-sense disambiguation, 
semantic role labeling, named entity extraction, 
and anaphora resolution). Text corpora have been
manually annotated with such data structures
in order to compare the performance of various 
systems. The availability of standard benchmarks has 
stimulated research in Natural Language Processing (NLP)
and effective systems have been designed for all these tasks.
Such systems are often viewed as software components
for constructing real-world NLP solutions.

The overwhelming majority of these state-of-the-art 
systems address a benchmark task by applying linear 
statistical models to ad-hoc features.  
In other words, the researchers themselves 
discover intermediate representations by 
engineering task-specific features. 
These features are often derived 
from the output of preexisting systems, 
leading to complex runtime dependencies. 
This approach is effective because researchers leverage a 
large body of linguistic knowledge.  On the other hand, 
there is a great temptation to optimize the performance 
of a system for a  specific benchmark.
Although such performance improvements can be very useful
in practice, they teach us little about the means to 
progress toward the broader goals of natural language understanding and the 
elusive goals of Artificial Intelligence.

In this contribution, we try to excel on \emph{multiple benchmarks} 
while \emph{avoiding task-specific enginering}.
Instead we use a \emph{single learning system} able to
discover adequate internal representations. 
In fact we view the benchmarks as indirect measurements of the 
relevance of the internal representations discovered 
by the learning procedure, and we posit that these intermediate 
representations are more general than any of the benchmarks.
Our desire to avoid task-specific engineered features 
led us to ignore a large body of linguistic knowledge.
Instead we reach good performance  
levels in most of the tasks by transferring intermediate 
representations discovered on large unlabeled datasets. 
We call this approach ``almost from scratch'' to emphasize the reduced 
(but still important) reliance on a priori NLP knowledge.

The paper is organized as follows.
\voirsec{sec-benchmark-tasks} 
describes the benchmark tasks of interest.
\voirsec{sec-general-nlp-nn} 
describes the unified model and reports benchmark 
results obtained with supervised training.
\voirsec{sec-lm} 
leverages large unlabeled datasets ($\sim $ $852$ million words)
to train the model on a language modeling task.
Performance improvements are then demonstrated
by transferring the unsupervised internal representations
into the supervised benchmark models.
\voirsec{sec-multi-task} 
investigates multitask supervised training.
\voirsec{sec-nlp-not-from-scratch} then
evaluates how much further improvement
can be achieved by incorporating standard NLP 
task-specific engineering into our systems.
Drifting away from our initial goals gives
us the opportunity to construct an all-purpose
tagger that is simultaneously accurate, 
 practical, and  fast.
We then conclude with a short discussion section.


\section{The Benchmark Tasks}
\label{sec-benchmark-tasks}

In this section, we briefly introduce four standard NLP tasks on which we
will benchmark our architectures within this paper: Part-Of-Speech tagging
(POS), chunking (CHUNK), Named Entity Recognition (NER) and Semantic Role
Labeling (SRL). For each of them, we consider a standard experimental
setup and give an overview of state-of-the-art systems on this setup. The
experimental setups are summarized in~\voirtbl{tbl-experimental-setup},
while state-of-the-art systems are reported in~\voirtbl{tbl-state-of-the-art}.
\begin{table}[h]
\begin{small}
\begin{tabular}{cccccc}
\textbf{Task} & \textbf{Benchmark} & \textbf{Dataset} & \textbf{Training set} & \textbf{Test set} & ~ \\
& & & (\#tokens) & (\#tokens) & (\#tags) \\ \hline
POS & \citet{toutanova:2003} & WSJ & sections 0--18 & sections 22--24 & (~45~) \\
& & & (~912,344~) & (~129,654~) &   \\ \hline
Chunking & CoNLL~2000 & WSJ & sections 15--18  & section 20 & (~42~)\\
& & & (~211,727~) & (~47,377~) & (IOBES) \\ \hline
NER & CoNLL~2003 & Reuters & ``eng.train'' & ``eng.testb'' & (~17~)\\
& & & (~203,621~) & (~46,435~) & (IOBES)\\ \hline
SRL & CoNLL~2005 & WSJ & sections 2--21 & section 23 & (~186~) \\
& & & (~950,028~) & + 3 Brown sections & (IOBES) \\
& & & & (~63,843~)  \\
\end{tabular}
\caption{\label{tbl-experimental-setup} Experimental setup: for each task, we report the standard benchmark we used,
the dataset it relates to, as well as training and test information.}
\end{small}
\end{table}
\begin{table}
\centering
\subfloat[POS] {
  \begin{minipage}{0.5\linewidth}
  \begin{tabular}{lc}
    System & Accuracy \\ \hline
    \citet{shen:2007} & 97.33\% \\
    \textbf{\citet{toutanova:2003}} & {97.24\%} \\
    \citet{gimenez:2004} & 97.16\%
  \end{tabular}
  \end{minipage}
}
\subfloat[CHUNK] {
  \begin{minipage}{0.5\linewidth}
  \begin{tabular}{lc}
    System & F1 \\ \hline
    \citet{shen:2005} & 95.23\% \\
    \textbf{\citet{sha:2003}} & {94.29\%} \\
    \citet{kudoh:2001} & 93.91\%
  \end{tabular}
  \end{minipage}
}\\
\subfloat[NER] {
  \begin{minipage}{0.5\linewidth}
  \begin{tabular}{lc}
    System & F1 \\ \hline
    \textbf{\citet{ando:2005}} & {89.31\%} \\
    \citet{florian:2003} & 88.76\% \\
    \citet{kudoh:2001} & 88.31\%
  \end{tabular}
  \end{minipage}
}
\subfloat[SRL] {
  \begin{minipage}{0.5\linewidth}
  \begin{tabular}{lc}
    System & F1 \\ \hline
    \textbf{\citet{koomen:2005}} & {77.92\%} \\
    \citet{pradhan:2005} & 77.30\% \\
    \citet{haghighi:2005} & 77.04\%
  \end{tabular}
  \end{minipage}
}
\caption{\label{tbl-state-of-the-art}
State-of-the-art systems on four NLP tasks. Performance is reported in
per-word accuracy for POS, and F1 score for CHUNK, NER and SRL. Systems in
bold will be referred as \emph{benchmark systems} in the rest of the paper
(see text).
}
\end{table}

\subsection{Part-Of-Speech Tagging}
POS aims at labeling each word with a unique tag that indicates its
\emph{syntactic role}, \eg.  plural noun, adverb, \dots
 A standard benchmark setup is described in detail
by~\citet{toutanova:2003}. Sections 0--18 of Wall Street Journal (WSJ)
 data are used for training,
while sections 19--21 are for validation and sections 22--24 for testing.

The best POS classifiers are based on classifiers trained on windows of text,
which are then fed to a bidirectional decoding algorithm during
inference. Features include preceding and following tag context as well as
multiple words (bigrams, trigrams\dots) context, and handcrafted
features to deal with unknown words.
\citet{toutanova:2003}, who use maximum entropy classifiers, and
a bidirectional dependency network~\citep{heckerman:2001} at inference,
reach $97.24\%$ per-word accuracy. \citet{gimenez:2004} proposed a SVM
approach also trained on text windows, with bidirectional inference
achieved with two Viterbi decoders (left-to-right and right-to-left). They
obtained $97.16\%$ per-word accuracy. More recently, \citet{shen:2007}
pushed the state-of-the-art up to $97.33\%$, with a new learning algorithm
they call \emph{guided learning}, also for bidirectional sequence
classification.



\subsection{Chunking}

Also called shallow parsing, chunking aims at labeling segments of a
sentence with syntactic constituents such as noun or verb phrases (NP or
VP). Each word is assigned only one unique tag, often encoded as a
begin-chunk (\eg. B-NP) or inside-chunk tag (\eg. I-NP). Chunking is often
evaluated using the CoNLL~2000 shared
task\footnote{See \url{http://www.cnts.ua.ac.be/conll2000/chunking}.}. Sections
15--18 of WSJ data are used for training and section 20 for
testing. Validation is achieved by splitting the training set.

\citet{kudoh:2000} won the CoNLL 2000 challenge on chunking with a F1-score of $93.48\%$.
Their system was based on Support Vector Machines (SVMs). Each SVM was
trained in a pairwise classification manner, and fed with a window around
the word of interest containing POS and words as features, as well as
surrounding tags. They perform dynamic programming at test time. 
Later, they improved their results up to $93.91\%$~\citep{kudoh:2001}
using an ensemble of classifiers trained with 
different tagging conventions (see~\voirsec{sec-tags}).

Since then, a certain number of systems based on second-order random fields
were reported~\citep{sha:2003,mcdonald:2005,sun:2008}, all reporting around
$94.3\%$ F1 score. These systems use features composed of words, POS
tags, and tags.

More recently, \citet{shen:2005} obtained $95.23\%$ using a voting
classifier scheme, where each classifier is trained on different tag
representations\footnote{See~\voirtbl{tbl-tagging-schemes} for tagging scheme details.}
(IOB, IOE, \dots). They use POS features coming from an
external tagger, as well carefully hand-crafted \emph{specialization}
features which again change the data representation by concatenating some
(carefully chosen) chunk tags or some words with their POS
representation. They then build trigrams over these features, which are
finally passed through a Viterbi decoder a test time.

\subsection{Named Entity Recognition}

NER labels atomic elements in the sentence into categories such as
``PERSON''  or ``LOCATION''. As in the chunking task, each word is
assigned a tag prefixed by an indicator of the beginning or the inside of
an entity. The CoNLL~2003
setup\footnote{See \url{http://www.cnts.ua.ac.be/conll2003/ner}.} is a 
NER benchmark dataset based on Reuters data. The contest provides training,
validation and testing sets.

\citet{florian:2003} presented the best system
at the NER CoNLL 2003 challenge, with $88.76\%$ F1 score. They used a
combination of various machine-learning classifiers. Features they picked
included words, POS tags, CHUNK tags, prefixes and suffixes, a large
gazetteer (not provided by the challenge), as well as the output of two
other NER classifiers trained on richer datasets. \citet{chieu:2003}, the
second best performer of CoNLL 2003 ($88.31\%$ F1), also used an external
gazetteer (their performance goes down to $86.84\%$ with no gazetteer) and
 several hand-chosen features.

Later, \citet{ando:2005} reached $89.31\%$ F1 with a semi-supervised
approach. They trained jointly a linear model on NER with a linear model on
two auxiliary unsupervised tasks. They also performed Viterbi decoding at
test time. The unlabeled corpus was 27M words taken from
Reuters. Features included words, POS tags, suffixes and prefixes or CHUNK
tags, but overall were less specialized than CoNLL 2003 challengers.

\subsection{Semantic Role Labeling}

SRL aims at giving a semantic role to a syntactic constituent of a
sentence. In the PropBank~\citep{propbank} formalism one assigns roles
ARG$0$-$5$ to words that are arguments of a verb (or more technically,
a \emph{predicate}) in the sentence, \eg. the following sentence might
be tagged ``[John]$_{ARG0}$ [ate]$_{REL}$ [the apple]$_{ARG1}$ '', where
``ate'' is the predicate. The precise arguments depend on a verb's {\em
frame} and if there are multiple verbs in a sentence some words might have
multiple tags. In addition to the ARG$0$-$5$ tags, there there are several
modifier tags such as ARGM-LOC (locational) and ARGM-TMP (temporal) that
operate in a similar way for all verbs. We picked
CoNLL~2005\footnote{See \url{http://www.lsi.upc.edu/~srlconll}.} as our SRL
benchmark. It takes sections 2--21 of WSJ data as training set, and section
24 as validation set. A test set composed of section 23 of WSJ concatenated
with 3 sections from the Brown corpus is also provided by the challenge.

State-of-the-art SRL systems consist of several stages: producing a parse
tree, identifying which parse tree nodes represent the arguments of 
a given verb, and finally classifying these nodes to compute 
the corresponding SRL tags. This entails extracting numerous
base features from the parse tree and feeding them into statistical models.
Feature categories commonly used by these system 
include~\citep{gildea,pradhan2004ssp}:
 \begin{itemize}\parskip=0pt
\item the parts of speech and syntactic labels of words and nodes in the tree;
\item the node's position (left or right) in relation to the verb;
\item the syntactic path to the verb in the parse tree;
\item whether a node in the parse tree is part of a noun or verb phrase;
\item the voice of the sentence: active or passive;
\item the node's head word; and
\item the verb sub-categorization.
 \end{itemize}

\citet{pradhan2004ssp} take these base features 
and define additional features, notably
the part-of-speech tag of the head word,
the predicted named entity class of the argument,
features providing word sense disambiguation for the verb
(they add 25 variants of 12 new feature types overall).
This system is close to the state-of-the-art in performance.
 \citet{pradhan:2005} obtain $77.30\%$ F1 with a system based on SVM
classifiers and simultaneously using the two parse 
trees provided for the SRL task.
In the same spirit, \citet{haghighi:2005} use 
log-linear models on each tree node, re-ranked globally with a
dynamic algorithm. Their system reaches $77.04\%$ using the 
five top Charniak parse trees.

 \citet{koomen:2005} hold the state-of-the-art 
with Winnow-like~\citep{littlestone:1998} classifiers, followed by a decoding
stage based on an integer program that enforces specific constraints on
SRL tags. They reach $77.92\%$ F1 on CoNLL 2005, thanks to the five top
parse trees produced by the \citet{charniak:2000} parser (only the first one was
provided by the contest) as well as the \citet{collins:1999} parse
tree.

\subsection{Evaluation}
\label{sec-evaluation}

In our experiments, we strictly followed the standard evaluation
procedure of each CoNLL challenges for NER, CHUNK and SRL. All these three
tasks are evaluated by computing the F1 scores over \emph{chunks} produced
by our models. The POS task is evaluated by computing the \emph{per-word}
accuracy, as it is the case for the standard benchmark we refer
to~\citep{toutanova:2003}. We picked the \texttt{conlleval}
script\footnote{Available at
  \url{http://www.cnts.ua.ac.be/conll2000/chunking/conlleval.txt}.} for
evaluating POS\footnote{We used the ``\texttt{-r}'' option of the
  \texttt{conlleval} script to get the per-word accuracy, for POS only.},
NER and CHUNK. For SRL, we used the \texttt{srl-eval.pl} script included in
the \texttt{srlconll} package\footnote{Available at
  \url{http://www.lsi.upc.es/~srlconll/srlconll-1.1.tgz}.}.

\subsection{Discussion}

When participating in an (open) challenge, it is legitimate to increase
generalization by all means. It is thus not surprising to see many top
CoNLL systems using \emph{external labeled data}, like additional NER
classifiers for the NER architecture of~\citet{florian:2003} or additional
parse trees for SRL systems~\citep{koomen:2005}. Combining multiple
systems or tweaking carefully features is also a common approach, like in
the chunking top system~\citep{shen:2005}.

However, when \emph{comparing} systems, we do not learn anything of the
quality of each system if they were trained with \emph{different} labeled
data. 
For that reason, we will refer
to \emph{benchmark systems}, that is, top existing systems which avoid usage of
external data and have been well-established in the NLP
field: \citep{toutanova:2003} for POS and \citep{sha:2003} for
chunking. For NER we consider \citep{ando:2005} as they were using
additional \emph{unlabeled} data only. We picked \citep{koomen:2005} for SRL, keeping
in mind they use 4 additional parse trees not provided by the challenge.
These benchmark systems will serve as baseline references in our
experiments. We marked them in bold in~\voirtbl{tbl-state-of-the-art}.

We note that for the four tasks we are considering in this work,
it can be seen that for the more complex tasks (with corresponding lower accuracies), 
the best systems proposed have more
engineered features relative to the best systems on the simpler tasks.
That is, the POS task is one of the simplest of our four tasks, and only has relatively
few engineered features, whereas SRL is the most complex, and many kinds of features
have been designed for it. This clearly has implications for as yet unsolved NLP tasks
requiring more sophisticated semantic understanding than the ones considered here.


\section{The Networks}
\label{sec-general-nlp-nn}

All the NLP tasks above can be seen as tasks assigning 
labels to words.
The traditional NLP approach is:  extract from the
sentence a rich set of hand-designed features  which
are then fed to a standard classification algorithm, \eg.
a Support Vector Machine (SVM), often with a linear kernel.
The choice of features is a
completely empirical process, mainly based first 
on linguistic intuition, and then trial and error,
and the feature selection is task dependent, 
implying additional research for each new NLP task. 
Complex tasks like SRL then require a large number of possibly
complex features (\eg., extracted from a parse tree) which
can impact the computational cost which might be important
for large-scale applications or applications requiring real-time response. 

Instead, we advocate  
a radically different approach: as input we will try to pre-process our features
as little as possible and then
use a multilayer neural network (NN) architecture, trained in an
 end-to-end fashion. The architecture takes the input sentence
and learns several layers of feature extraction that process the inputs.
The features computed by the deep layers of the network
are automatically trained by backpropagation to be relevant to the task.
 We describe in this section
a general multilayer architecture suitable for all our NLP tasks, which is 
generalizable to other NLP tasks as well.

\begin{figure}[tb]
\centering
\includegraphics[scale=1]{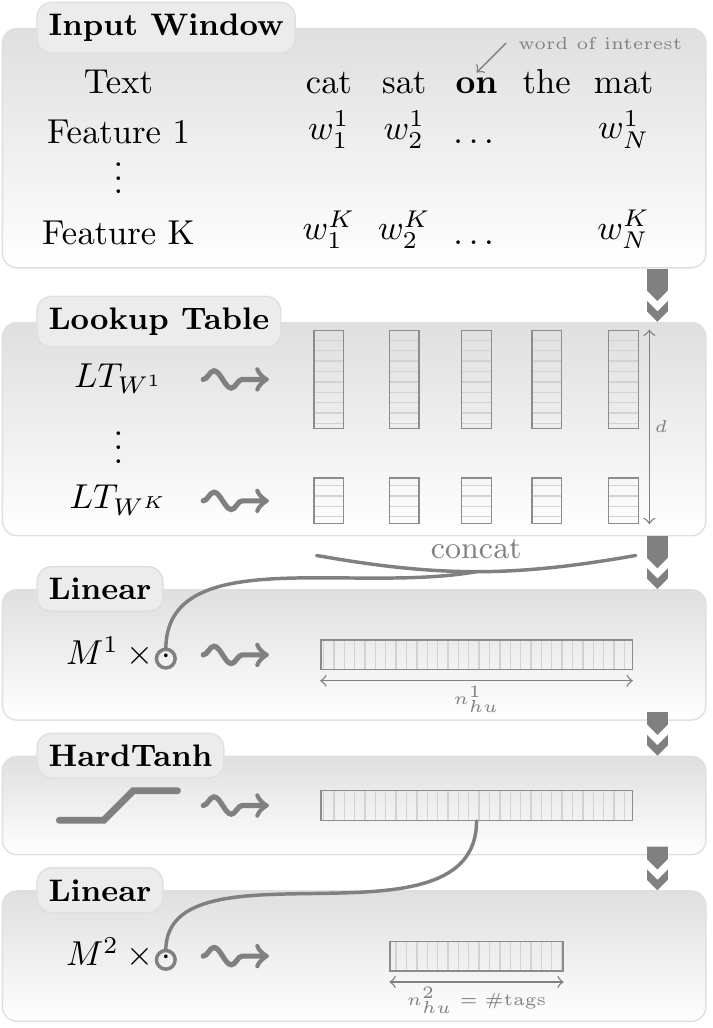}
\caption{\label{fig-net-window} Window approach network.}
\end{figure}

\begin{figure}[tb]
\centering
\includegraphics[scale=1]{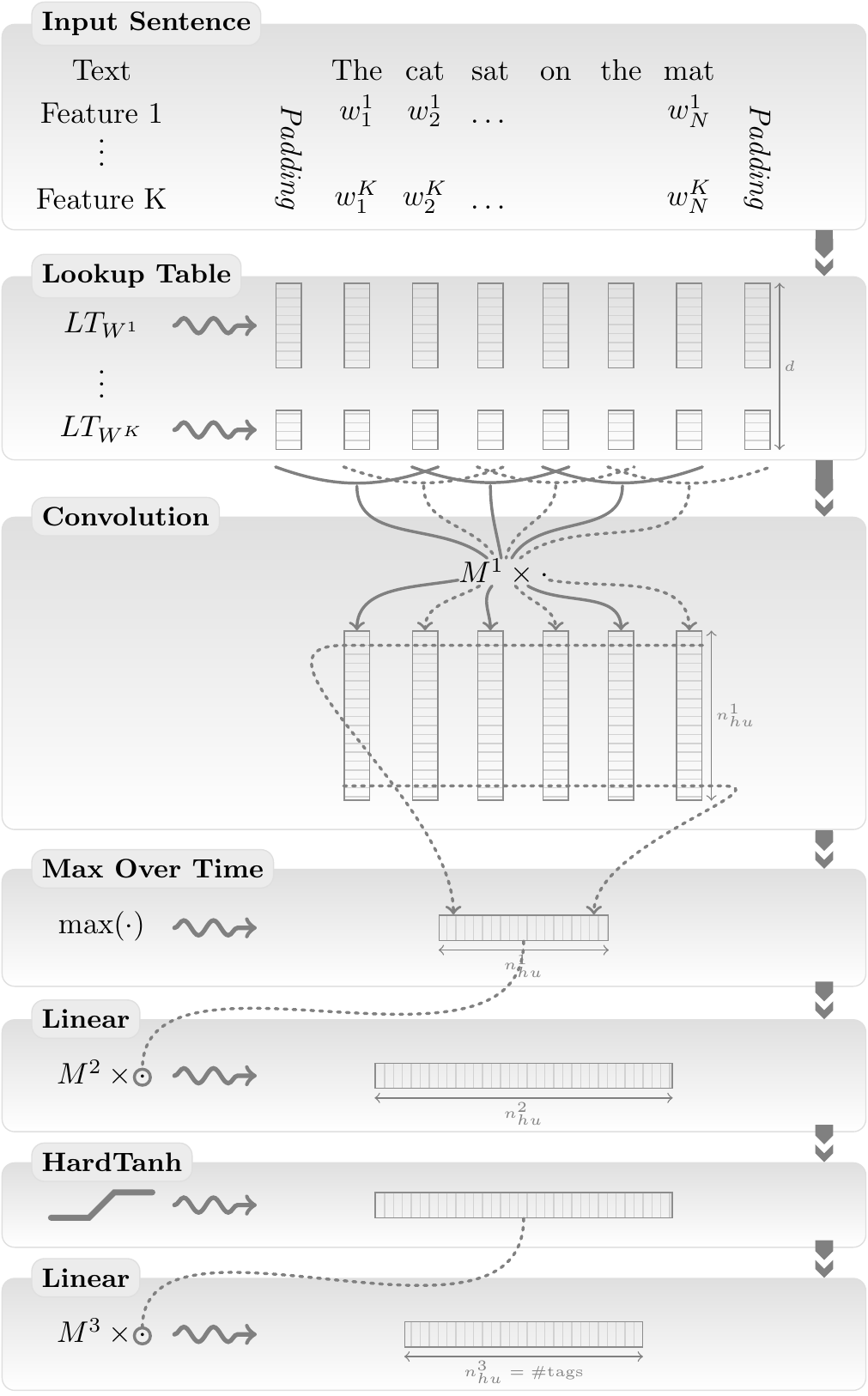}
\caption{\label{fig-net-sentence} Sentence approach network.}
\end{figure}

Our architecture is summarized in~\voirfig{fig-net-window} 
and~\voirfig{fig-net-sentence}.
The first layer extracts features for each word.
The second layer extracts features 
from a window of words or from the whole sentence, 
treating it as a {\em sequence} with local and global
structure (i.e., it is not treated like a bag of words).
The following layers are standard NN layers.

\paragraph{Notations}

We consider a neural network $f_{\bm{\theta}}(\cdot)$, with parameters
$\bm{\theta}$.  Any feed-forward neural network with $L$ layers, can be
seen as a composition of functions $f^l_{\bm{\theta}}(\cdot)$,
corresponding to each layer~$l$:
\begin{equation*}
  f_{\bm{\theta}}(\cdot) = f^L_{\bm{\theta}}(f^{L-1}_{\bm{\theta}}(\ldots f^1_{\bm{\theta}}(\cdot) \ldots))\,.
\end{equation*}
In the following, we will describe each layer we use in our networks shown
in~\voirfig{fig-net-window} and~\voirfig{fig-net-sentence}. We adopt few
notations. Given a matrix $A$ we denote $\midx{A}{i}{j}$ the coefficient at
row $i$ and column $j$ in the matrix. We also denote
$\window{A}{i}{d_{win}}$ the vector obtained by concatenating the $d_{win}$
column vectors around the $i^{th}$ column vector of matrix $A\in\R^{d_1\times d_2}$:
\begin{equation*}
\left[ \window{A}{i}{d_{win}} \right]^{\T} = \left(\midx{A}{1}{i-d_{win}/2}\,\ldots\,\midx{A}{d_1}{i-d_{win}/2},\,\ldots,\,
  \midx{A}{1}{i+d_{win}/2}\,\ldots\,\midx{A}{d_1}{i+d_{win}/2} \right)\,.
\end{equation*}
As a special case, $\window{A}{i}{1}$ represents the $i^{th}$ column of
matrix $A$.  For a vector $v$, we denote $\vidx{v}{i}$ the scalar at index
$i$ in the vector. Finally, a sequence of element $\{x_1,\,x_2,\,\ldots,\,
x_T\}$ is written $\seq{x}{1}{T}$. The $i^{th}$ element of the sequence is
$\seq{x}{i}{}$.


\subsection{Transforming Words into Feature Vectors}
\label{sec-words-features}

One of the essential key points of our architecture is its ability to perform
well with the use of (almost\footnote{We did some pre-processing, namely lowercasing and encoding capitalization as another feature. With enough (unlabeled) training data, presumably we could learn a model without this processing. Ideally, an even more raw input would be to learn 
from letter sequences rather than words, however we felt that this was beyond the scope of this work.}) raw words.
The ability for our method to learn good word representations is thus 
crucial to our approach. For efficiency,
words are fed to our architecture as indices taken from a finite dictionary
${\cal D}$. Obviously, a simple index does not carry much useful
information about the word. However, the first layer of our network maps
each of these word indices into a feature vector, by a lookup table
operation. Given a task of interest, a relevant representation of
each word is then given by the corresponding lookup table feature vector,
which is trained by backpropagation.

More formally, for each word $w\in {\cal D}$, an internal $d_{wrd}$-dimensional
feature vector representation is given by the \emph{lookup table} layer
$LT_{W}(\cdot)$:
\begin{displaymath}
LT_{W}(w) = \col{W}{w}\,,
\end{displaymath}
where $W\in\R^{d_{wrd}\times|{\cal D}|}$ is a matrix of parameters to be {learnt},
$\col{W}{w} \in \R^{d_{wrd}}$ is the $w^{th}$ column of $W$ and $d_{wrd}$ is the word vector
size (a hyper-parameter to be chosen by the user). Given a sentence or any sequence
of $T$ words $\seq{w}{1}{T}$ in ${\cal D}$, the lookup table layer
applies the same operation for each word in the sequence, producing the following
output matrix:
\begin{equation}
\label{eq-lookup-table-layer}
LT_{W}(\seq{w}{1}{T}) = \left( \begin{array}{cccc} \col{W}{\seq{w}{1}{}} & \col{W}{\seq{w}{2}{}} & \ldots & \col{W}{\seq{w}{T}{}} \end{array}\right)\,.
\end{equation}
This matrix can then be fed to further neural network layers, as we
will see below.

\subsubsection{Extending to Any Discrete Features}
\label{sec-discrete-features}

One might want to provide features other than words if one suspects that these features are
 helpful for the task of interest. For example, for the NER
task, one could provide a feature which says if a word is in a gazetteer or
not. Another common practice is to introduce some basic pre-processing,
such as word-stemming or dealing with upper and lower case. In this latter
option, the word would be then represented by three discrete features: its
lower case stemmed root, its lower case ending, and a capitalization
feature.

Generally speaking, we can consider a word as represented by $K$ discrete
features $\bm{w} \in {\cal D}^1\times \dots
\times {\cal D}^K$, where ${\cal D}^{k}$ is the dictionary for the $k^{th}$
feature. We associate to each feature a lookup table $LT_{W^k}(\cdot)$,
with parameters $W^k\in\R^{d^k_{wrd}\times|{\cal D}^k|}$ where $d^k_{wrd} \in \N$ is a
user-specified vector size. Given a word $\bm{w}$, a feature vector of dimension $d_{wrd}=\sum_k
d^k_{wrd}$ is then obtained by concatenating all lookup table outputs:
\begin{displaymath}
  LT_{W^1,\dots,W^K}(\bm{w})
  = \left( \begin{array}{c} LT_{W^1}(w_1) \\
                            \vdots \\
                            LT_{W^K}(w_K) \end{array} \right)
  = \left( \begin{array}{c} \col{W^1}{w_1} \\
                            \vdots \\
                            \col{W^K}{w_K} \end{array} \right)\,.
\end{displaymath}
The matrix output of the lookup table layer for a sequence of words
$\seq{\bm{w}}{1}{T}$ is then similar to~\voireq{eq-lookup-table-layer}, but
where extra rows have been added for each discrete feature:
\begin{equation}
\label{eq-lookup-table-multi}
  LT_{W^1,\dots,W^K}(\seq{\bm{w}}{1}{T})
  = \left( \begin{array}{ccc} \col{W^1}{\seq{w_1}{1}{}} & \ldots & \col{W^1}{\seq{w_1}{T}{}} \\
                              \vdots      &        & \vdots \\
                              \col{W^K}{\seq{w_K}{1}{}} & \ldots & \col{W^K}{\seq{w_K}{T}{}} \end{array} \right)\,.
\end{equation}
These vector features in the lookup table effectively learn features for words in the dictionary. 
Now, we want to use these trainable features as input to further layers of trainable feature extractors, that can represent groups of words and then finally sentences. 

\subsection{Extracting Higher Level Features from Word Feature Vectors}

Feature vectors produced by the lookup table layer need to be combined in
subsequent layers of the neural network to produce a tag decision for each
word in the sentence. Producing tags for each element in variable length
sequences (here, a sentence is a sequence of words) is a standard problem
in machine-learning.  We consider two common approaches which tag \emph{one
  word at the time}: a window approach, and a (convolutional) sentence approach.

\subsubsection{Window Approach}
\label{sec-window-approach}

A window approach assumes the tag of a word depends mainly on its
neighboring words. Given a word to tag, we consider a fixed size $k_{sz}$
(a hyper-parameter) window of words around this word. Each word in the window
is first passed through the lookup table
layer~\voireq{eq-lookup-table-layer} or~\voireq{eq-lookup-table-multi},
producing a matrix of word features of fixed size $d_{wrd} \times k_{sz}$. This
matrix can be viewed as a $d_{wrd}\, k_{sz}$-dimensional vector by concatenating
each column vector, which can be fed to further neural network layers.
More formally, the word feature window given by the first network layer can be written as:
\begin{equation}
\label{eq-word-feature-window}
\f{1} = \window{LT_W(\seq{w}{1}{T})}{t}{d_{win}} = \left( \begin{array}{c}
                                                         \col{W}{\seq{w}{t-d_{win}/2}{}} \\
                                                         \vdots \\
                                                         \col{W}{\seq{w}{t}{}} \\
                                                         \vdots \\
                                                         \col{W}{\seq{w}{t+d_{win}/2}{}} \\
                                                       \end{array}
                                                \right)\,.
\end{equation}
\paragraph{Linear Layer}  The fixed size vector $\f{1}$ can be fed to one or several standard
neural network layers which perform affine transformations over their
inputs:
\begin{equation}
\label{eq-linear-layer}
\f{l} = W^l\, \f{l-1} \, + \, \bm{b}^l\,,
\end{equation}
where $W^l \in \R^{n_{hu}^l\times n_{hu}^{l-1}}$ and $\bm{b}^l \in
\R^{n_{hu}^l}$ are the parameters to be \emph{trained}. The hyper-parameter
$n_{hu}^l$ is usually called the \emph{number of hidden units} of the
$l^{th}$ layer.

\paragraph{HardTanh Layer} Several linear layers are often stacked, interleaved
with a non-linearity function, to extract highly non-linear features. If no
non-linearity is introduced, our network would be a simple linear model. We
chose a ``hard'' version of the hyperbolic tangent as non-linearity.
It has the advantage of being slightly cheaper to compute
(compared to the exact hyperbolic tangent), while leaving the generalization
performance unchanged~\citep{collobert:2004}. The corresponding layer $l$
applies a HardTanh over its input vector:
\begin{equation*}
\vidx{\f{l}}{i} = \textrm{HardTanh}(\vidx{\f{l-1}}{i})\,,
\end{equation*}
where
\begin{equation}
\label{eq-hardtanh-layer}
\textrm{HardTanh}(x) = \left\{ \begin{array}{rl}
                                            -1 & \textrm{if} \ x < -1 \\
                                             x & \textrm{if} \ -1 <= x <= 1 \\
                                             1 & \textrm{if} \ x > 1
                               \end{array}
\right.\,.
\end{equation}

\paragraph{Scoring} Finally, the output size of the last layer $L$ of our network is equal to
the number of possible tags for the task of interest. Each output can be
then interpreted as a \emph{score} of the corresponding tag (given the
input of the network), thanks to a carefully chosen cost function that we
will describe later in this section.

\begin{remark}[Border Effects]
The feature window~\voireq{eq-word-feature-window} is not well defined for
words near the beginning or the end of a sentence. To circumvent this
problem, we augment the sentence with a special ``PADDING'' word
replicated $d_{win}/2$ times at the beginning and the end. 
This is akin to the use of ``start'' and ``stop''
symbols in sequence models. 
\end{remark}

\subsubsection{Sentence Approach}
\label{sec-sentence-approach}

We will see in the experimental section that a window approach performs
well for most natural language processing tasks we are interested
in. However this approach fails with SRL, where the tag of a word depends
on a verb (or, more correctly, predicate) chosen beforehand in the sentence.
 If the verb 
 falls outside the
window, one cannot expect this word to be tagged correctly. In this
particular case, tagging a word requires the consideration of the
\emph{whole} sentence. When using neural networks, the natural choice to
tackle this problem becomes a convolutional approach, first introduced by
\citet{waibel:1989} and also called Time Delay Neural Networks (TDNNs) in
the literature.

We describe in detail our convolutional network below. It successively
takes the complete sentence, passes it through the lookup table
layer~\voireq{eq-lookup-table-layer}, produces local features around each
word of the sentence thanks to convolutional layers, combines these feature
into a global feature vector which can then be fed to standard affine
layers~\voireq{eq-linear-layer}. In the semantic role labeling case, this
operation is performed for each word in the sentence, and for each verb in
the sentence. It is thus necessary to encode in the network architecture
which verb we are considering in the sentence, and which word we want to
tag. For that purpose, each word at position $i$ in the sentence is
augmented with two features in the way described
in~\voirsec{sec-discrete-features}. These features encode the relative
distances $i-pos_v$ and $i-pos_w$ with respect to the chosen verb at
position $pos_v$, and the word to tag at position $pos_w$ respectively.

\paragraph{Convolutional Layer }
A convolutional layer can be seen as a generalization of a window approach:
given a sequence represented by columns in a matrix $\f{l-1}$ (in our
lookup table matrix~\voireq{eq-lookup-table-layer}), a matrix-vector
operation as in~\voireq{eq-linear-layer} is applied to each window of
successive windows in the sequence. Using previous notations, the
$t^{th}$~output column of the $l^{th}$~layer can be computed as:
\begin{equation}
\label{eq-convolution-layer}
\col{\f{l}}{t} = W^l\, \window{\f{l-1}}{t}{d_{win}} + \bm{b}^l \ \ \ \ \forall t\,,
\end{equation}
where the weight matrix $W^l$ is the same across all windows $t$ in the
sequence.  Convolutional layers extract local features around each window
of the given sequence.  As for standard affine
layers~\voireq{eq-linear-layer}, convolutional layers are often stacked to
extract higher level features. In this case, each layer must be followed by
a non-linearity~\voireq{eq-hardtanh-layer} or the network would be
equivalent to one convolutional layer.

\begin{figure}[tb]
\centering
\includegraphics[width=0.49\linewidth]{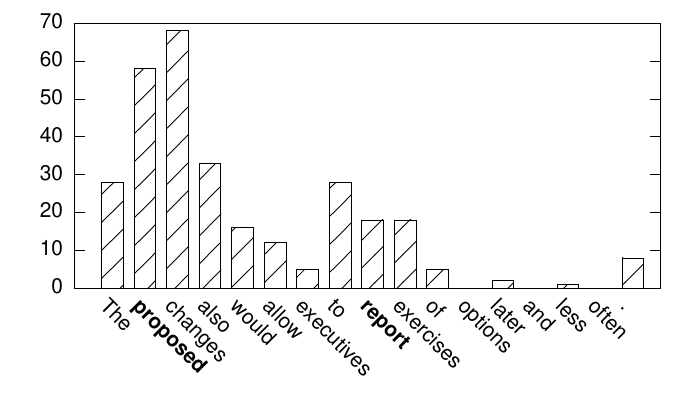}
\includegraphics[width=0.49\linewidth]{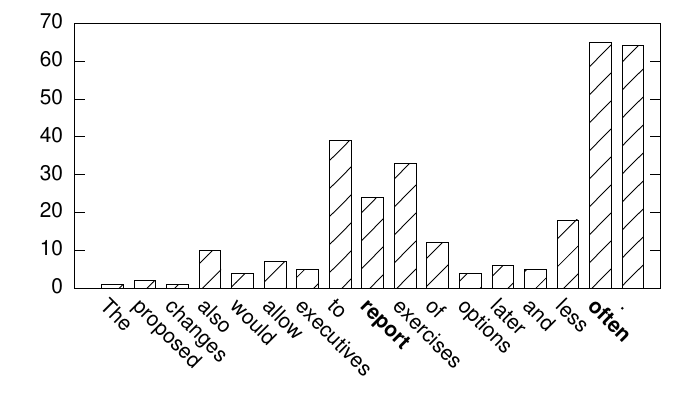}
\caption{\label{fig-srl-max} Number of features chosen at each word
position by the Max layer.
We consider a sentence approach network (\voirfig{fig-net-sentence})
trained for SRL. The number of ``local'' features output by the convolution
layer is $300$ \emph{per word}. By applying a Max over the sentence, we
obtain $300$ features for the \emph{whole sentence}. It is interesting to
see that the network catches features mostly around the verb of interest
(here ``report'') and word of interest (``proposed''~(left) or
``often''~(right)).
}
\end{figure}

\paragraph{Max Layer}
The size of the output~\voireq{eq-convolution-layer} depends on the number
of words in the sentence fed to the network. Local feature vectors
extracted by the convolutional layers have to be combined to obtain a
global feature vector, with a fixed size independent of the sentence
length, in order to apply subsequent standard affine layers.  Traditional
convolutional networks often apply an average (possibly weighted) or a max
operation over the ``time'' $t$ of the
sequence~\voireq{eq-convolution-layer}. (Here, ``time'' just means the
position in the sentence, this term stems from the use of convolutional
layers in e.g. speech data where the sequence occurs over time.)  The
average operation does not make much sense in our case, as in general most
words in the sentence do not have any influence on the semantic role of a
given word to tag.
Instead, we used a max approach, which forces the network to capture the
most useful local features produced by the convolutional layers
(see~\voirfig{fig-srl-max}), for the task at hand. Given a \emph{matrix}
$\f{l-1}$ output by a convolutional layer $l-1$, the Max layer $l$ outputs
a \emph{vector} $\f{l}$:
\begin{equation}
\label{eq-max-layer}
\vidx{\f{l}}{i} = \max_{t} \midx{\f{l-1}}{i}{t}\ \ \ \ \ \ 1 \leq i \leq n_{hu}^{l-1}\,.
\end{equation}
This fixed sized global feature vector can be then fed to standard affine
network layers~\voireq{eq-linear-layer}. As in the window approach, we then finally 
 produce one score per possible tag for the given task.

\begin{remark}
The same border effects arise in the convolution
operation~\voireq{eq-convolution-layer} as in the window
approach~\voireq{eq-word-feature-window}. We again work around 
this problem by padding the sentences with a special word. 
\end{remark}

\subsubsection{Tagging Schemes}
\label{sec-tags}

As explained earlier, the network output layers compute 
scores for all the possible tags for the task of interest.
In the window approach, these tags apply to the word 
located in the center of the window. In the (convolutional) sentence approach, 
these tags apply to the word designated by 
additional markers in the network input.

The POS task indeed consists of marking the syntactic role of
each word. However, the remaining three tasks associate labels
with segments of a sentence.  This is usually achieved by using 
special tagging schemes to identify the segment boundaries, as shown
in~\voirtbl{tbl-tagging-schemes}.
Several such schemes have been defined (IOB, IOE, IOBES, \dots)
without clear conclusion as to which scheme is better in general.
State-of-the-art performance is sometimes obtained by combining classifiers
trained with different tagging schemes~\citep[\eg.][]{kudoh:2001}.
\begin{table}[tb]
\begin{center}
\begin{tabular}{l|c|c|c|c|c}
Scheme & Begin  & Inside  & End  & Single  & Other \\
IOB    & B-X    & I-X     & I-X  & B-X     & O\\
IOE    & I-X    & I-X     & E-X  & E-X     & O\\
IOBES  & B-X    & I-X     & E-X  & S-X     & O\\
\end{tabular}
\caption{\label{tbl-tagging-schemes} Various tagging schemes. Each word in
  a segment labeled ``X'' is tagged with a prefixed label, depending of the
  word position in the segment (begin, inside, end). Single word segment
  labeling is also output. Words not in a labeled segment are labeled
  ``O''. Variants of the IOB (and IOE) scheme exist, where the prefix B (or E) is
  replaced by I for all segments not contiguous with another segment having
  the same label ``X''.
}
\end{center}
\end{table}

The ground truth for the NER, CHUNK, and SRL tasks is provided using two
different tagging schemes.  In order to eliminate this additional source of
variations, we have decided to use the most expressive IOBES tagging scheme
for all tasks.  For instance, in the CHUNK task, we describe noun phrases
using four different tags.  Tag ``S-NP'' is used to mark a noun phrase
containing a single word.  Otherwise tags ``B-NP'', ``I-NP'', and ``E-NP''
are used to mark the first, intermediate and last words of the noun phrase.
An additional tag ``O'' marks words that are not members of a chunk.
During testing, these tags are then converted to the original IOB tagging
scheme and fed to the standard performance evaluation scripts mentioned
in~\voirsec{sec-evaluation}.

\subsection{Training}

All our neural networks are trained by maximizing a likelihood over the
training data, using stochastic gradient ascent. If we denote $\bm{\theta}$ to be 
all the trainable parameters of the network, which are 
trained using a training set ${\cal T}$
we want to maximize the following log-likelihood with respect to $\bm{\theta}$:
\begin{equation}
\label{eq-log-likelihood}
  \bm{\theta} \mapsto \sum_{(\bm{x},\,y)\in {\cal T}} \log p(y \,|\, \bm{x},\, \bm{\theta})\,,
\end{equation}
where $\bm{x}$ corresponds to either a training word window or a sentence
and its associated features, and $y$ represents the corresponding tag. The
probability $p(\cdot)$ is computed from the outputs of the neural
network. We will see in this section two ways of interpreting neural
network outputs as probabilities.

\subsubsection{Word-Level Log-Likelihood}
\label{sec-isolated-tag-criterion}

In this approach, each word in a sentence is considered
independently. Given an input example $\bm{x}$, the network with parameters
$\bm{\theta}$ outputs a score $\vidx{\f{}(\bm{x})}{i}$, for the $i^{th}$
tag with respect to the task of interest. To simplify the notation, we drop
$\bm{x}$ from now, and we write instead $\vidx{\f{}}{i}$.  This score can
be interpreted as a conditional tag probability
$p(i\,|\,\bm{x},\,\bm{\theta})$ by applying a softmax~\citep{bridle:1990}
operation over all the tags:
\begin{equation}
\label{eq-word-softmax}
p(i\,|\,\bm{x},\bm{\theta}) = \frac{e^{\vidx{\f{}}{i}}}{\sum_j e^{\vidx{\f{}}{j}}}\,.
\end{equation}
Defining the log-add operation as
\begin{equation}
\label{eq-logadd}
\logadd_i z_i = \log(\sum_i e^{z_i})\,,
\end{equation}
we can express the log-likelihood for one training example $(\bm{x}, y)$ as follows:
\begin{equation}
\label{eq-word-likelihood}
  \log p(y \,|\, \bm{x},\,\bm{\theta}) = \vidx{\f{}}{y} - \logadd_j \vidx{\f{}}{j}\,.
\end{equation}
While this training criterion, often referred as \emph{cross-entropy} is
widely used for classification problems, it might not be ideal in our case,
where there is often a correlation between the tag of a word in a sentence
and its neighboring tags. We now describe another common approach for 
neural networks which enforces dependencies between the predicted tags in a sentence.

\subsubsection{Sentence-Level Log-Likelihood}

In tasks like chunking, NER or SRL we know that there are dependencies
between word tags in a sentence: not only are tags organized in chunks, but
some tags cannot follow other tags. Training using a word-level
approach discards this kind of labeling information. We consider a training scheme
which takes into account the sentence structure: given the predictions of 
\emph{all tags} by our network for \emph{all words} in a sentence,
and given a score for going from one tag to another tag, we want to
encourage valid paths of tags during training, while discouraging all
other paths.

We consider the \emph{matrix} of scores $\f{}(\seq{\bm{x}}{1}{T})$ output by the network.
As before, we drop the input $\seq{\bm{x}}{1}{T}$ for notation simplification. The element
$\midx{\f{}}{i}{t}$ of the matrix is the score
output by the network with parameters $\bm{\theta}$, for the sentence
$\seq{\bm{x}}{1}{T}$ and for the $i^{th}$ tag, at the $t^{th}$ word. We
introduce a transition score $\midx{A}{i}{j}$  for jumping from $i$ to $j$ tags in successive words,
 and
an initial score $\midx{A}{i}{0}$ for starting from the $i^{th}$ tag. As the transition
scores are going to be trained (as are all network parameters $\bm{\theta}$), we
define $\tilde{\bm{\theta}} = \bm{\theta} \cup \{\midx{A}{i}{j} \ \forall
i,j\}$. The score of a sentence $\seq{\bm{x}}{1}{T}$ along a path of tags
$\seq{i}{1}{T}$ is then given by the sum of transition scores and network
scores:
\begin{equation}
\label{eq-path-sentence-score}
s(\seq{\bm{x}}{1}{T},\, \seq{i}{1}{T},\, \tilde{\bm{\theta}}) = \sum_{t=1}^T \left( \midx{A}{\seq{i}{t-1}{}}{\seq{i}{t}{}} + \midx{\f{}}{\seq{i}{t}{}}{t} \right)\,.
\end{equation}
Exactly as for the word-level likelihood~\voireq{eq-word-likelihood}, where
we were normalizing with respect to all \emph{tags} using a
softmax~\voireq{eq-word-softmax}, we normalize this score
over all possible \emph{tag paths} $\seq{j}{1}{T}$ using a softmax,
and we interpret the resulting ratio as 
a conditional \emph{tag path} probability.
Taking the $\log$, the conditional probability 
of the true path $\seq{y}{1}{T}$ is therefore given by:
\begin{equation}
\label{eq-sentence-likelihood}
\log p(\seq{y}{1}{T} \,|\, \seq{\bm{x}}{1}{T},\, \tilde{\bm{\theta}}) = s(\seq{\bm{x}}{1}{T},\, \seq{y}{1}{T},\, \tilde{\bm{\theta}}) - \logadd_{\forall \seq{j}{1}{T}} s(\seq{\bm{x}}{1}{T},\, \seq{j}{1}{T},\, \tilde{\bm{\theta}})\,.
\end{equation}
While the number of terms in the $\logadd$
operation~\voireq{eq-word-likelihood} was equal to the number of tags, it
grows exponentially with the length of the sentence
in~\voireq{eq-sentence-likelihood}. Fortunately, one can compute it in
linear time with the following standard recursion over $t$, taking
advantage of the associativity and distributivity on the semi-ring\footnote{In other words, read $\logadd$ as $\oplus$ and $+$ as $\otimes$.} $(\R\cup\{-\infty\},\, \logadd,\, +)$:
\begin{equation}
\label{eq-forward-score-recursion}
\begin{split}
\delta_t(k) & \overset{\Delta}{=} \logadd_{\{\seq{j}{1}{t}\,\cap\, \seq{j}{t}{} = k\}}  s(\seq{\bm{x}}{1}{t},\, \seq{j}{1}{t},\, \tilde{\bm{\theta}}) \\
            & = \logadd_i \logadd_{\{\seq{j}{1}{t} \,\cap\, \seq{j}{t-1}{}=i \,\cap\, \seq{j}{t}{}=k\}} s(\seq{\bm{x}}{1}{t},\, \seq{j}{1}{t-1},\, \tilde{\bm{\theta}}) + \midx{A}{\seq{j}{t-1}{}}{k} + \midx{\f{}}{k}{t}  \\
            & = \logadd_i \delta_{t-1}(i) + \midx{A}{i}{k} + \midx{\f{}}{k}{t}  \\
            & = \midx{\f{}}{k}{t} + \logadd_{i}\, \left( \delta_{t-1}(i) + \midx{A}{i}{k} \right) \ \ \ \ \forall k\,,
\end{split}
\end{equation}
followed by the termination
\begin{equation}
\label{eq-forward-score-termination}
\logadd_{\forall \seq{j}{1}{T}} s(\seq{\bm{x}}{1}{T},\, \seq{j}{1}{T},\, \tilde{\bm{\theta}}) = \logadd_{i}\, \delta_T(i)\,.
\end{equation}
We can now maximize in~\voireq{eq-log-likelihood} the
log-likelihood~\voireq{eq-sentence-likelihood} over all the training pairs
$(\seq{\bm{x}}{1}{T},\, \seq{y}{1}{T})$.

At inference time, given a sentence $\seq{\bm{x}}{1}{T}$ to tag, we have to
find the best tag path which minimizes the sentence
score~\voireq{eq-path-sentence-score}. In other words, we must find
\begin{equation}
\label{eq-sentence-inference}
\argmax_{\seq{j}{1}{T}} s(\seq{\bm{x}}{1}{T},\, \seq{j}{1}{T},\, \tilde{\bm{\theta}})\,.
\end{equation}
The Viterbi algorithm is the natural choice for this inference. It
corresponds to performing the recursion~\voireq{eq-forward-score-recursion}
and~\voireq{eq-forward-score-termination}, but where the $\logadd$ is
replaced by a $\max$, and then tracking back the optimal path through
each $\max$.

\begin{remark}[Graph Transformer Networks]
Our approach is a particular case of the discriminative forward training for
graph transformer networks (GTNs) \citep{bottou-1997,lecun:1998b}. The
log-likelihood~\voireq{eq-sentence-likelihood} can be viewed as the
difference between the forward score constrained over the valid paths (in our
case there is only the labeled path) and the unconstrained forward
score~\voireq{eq-forward-score-termination}.
\end{remark}

\begin{remark}[Conditional Random Fields]
An important feature of equation \voireq{eq-path-sentence-score}
is the absence of normalization. Summing the exponentials 
$e^{\:\midx{\f{}}{i}{t}}$ 
over all possible tags does not necessarily yield the unity.
If this was the case, the scores could be viewed
as the logarithms of conditional transition probabilities, 
and our model would be subject to the label-bias problem that 
motivates Conditional Random Fields~(CRFs)~\citep{lafferty:2001}.
The denormalized scores should instead be likened to the
potential functions of a CRF. In fact, a CRF maximizes 
the same likelihood~\voireq{eq-sentence-likelihood}
using a linear model instead of a nonlinear neural network.
CRFs have been widely used in the NLP world, such as for POS
tagging~\citep{lafferty:2001}, chunking~\citep{sha:2003},
NER~\citep{mccallum:2003} or SRL~\citep{cohn:2005}. 
Compared to such CRFs, we take advantage of the nonlinear network 
to learn appropriate features for each task of interest.
\end{remark}

\subsubsection{Stochastic Gradient}

Maximizing~\voireq{eq-log-likelihood} with stochastic
gradient~\citep{bottou:1991} is achieved by iteratively selecting a random
example $(\bm{x},\,y)$ and making a gradient step:
\begin{equation}
\label{eq-criterion-derivative}
\bm{\theta} \longleftarrow \bm{\theta} + \lambda\, \frac{\partial \log p(y \,|\, \bm{x},\, \bm{\theta})}{\partial \bm{\theta}}\,,
\end{equation}
where $\lambda$ is a chosen learning rate. Our neural networks described
in~\voirfig{fig-net-window} and~\voirfig{fig-net-sentence} are a succession
of layers that correspond to successive composition of functions. The neural
network is finally composed with the word-level log-likelihood~\voireq{eq-word-likelihood},
or successively composed in the
recursion~\voireq{eq-forward-score-recursion} if using the sentence-level
log-likelihood~\voireq{eq-sentence-likelihood}. Thus, an
\emph{analytical} formulation of the
derivative~\voireq{eq-criterion-derivative} can be computed, by applying
the differentiation chain rule through the network, and through the
word-level log-likelihood~\voireq{eq-word-likelihood} or through the
recurrence~\voireq{eq-forward-score-recursion}.

\begin{remark}[Differentiability]
Our cost functions are differentiable almost everywhere.
Non-differentiable points arise because we use a ``hard'' transfer 
function~\voireq{eq-hardtanh-layer} and because we use a ``max'' 
layer~\voireq{eq-max-layer} in the sentence approach network.
Fortunately, stochastic gradient still converges
to a meaningful local minimum despite such minor
differentiability problems~\citep{bottou:1991,bottou-1998}.
Stochastic gradient iterations that hit a non-differentiability are simply
skipped.
\end{remark}

\begin{remark}[Modular Approach]
The well known ``back-propagation''
algorithm~\citep{lecun:1985,rumelhart:1986}
computes gradients using the chain rule.
The chain rule can also be used 
in a modular implementation.\footnote{See \url{http://torch5.sf.net}.}
Our modules correspond to the boxes 
in~\voirfig{fig-net-window}
and~\voirfig{fig-net-sentence}.
Given derivatives with respect to its outputs, 
each module can independently compute derivatives 
with respect to its inputs and with respect to its 
trainable parameters, as proposed by~\citet{bottou:1990}. 
This allows us to easily build variants of our networks. For details
about gradient computations, see~Appendix~\ref{apx-gradients}.
\end{remark}

\begin{remark}[Tricks\label{rmk-tricks}]
Many tricks have been reported for training neural
networks~\citep{lecun:1998}. Which ones to choose is often confusing.
We employed only two of them: the initialization and update of the
parameters of each network layer were done according to the ``fan-in'' of
the layer, that is the number of inputs used to compute each output of this
layer~\citep{plaut:1987}. The fan-in for the lookup 
table~\voireq{eq-lookup-table-layer}, the
$l^{\rm th}$ linear layer~\voireq{eq-linear-layer} and the convolution
layer~\voireq{eq-convolution-layer} are respectively $1$, $n_{hu}^{l-1}$
and $d_{win}\times n_{hu}^{l-1}$. The initial parameters of the network
were drawn from a centered uniform distribution, with a variance equal to
the inverse of the square-root of the fan-in. The learning rate
in~\voireq{eq-criterion-derivative} was divided by the
fan-in, but stays fixed during the training.
\end{remark}

\subsection{Supervised Benchmark Results}
\label{sec-supervised-results}
\label{before-semi-sup}

\begin{table}[tb]
\begin{center}
\begin{tabular}{l|c|c|c|c}
{\bf Approach}    & {\bf POS}  & {\bf Chunking} & {\bf NER} & {\bf SRL} \\
                & (PWA) & (F1) & (F1) & (F1) \\ \hline
\textbf{Benchmark Systems} & {97.24} & {94.29} & {89.31} & {77.92} \\ \hline
NN+WLL          & 96.31 & 89.13 & 79.53 & 55.40 \\
NN+SLL          & 96.37  & 90.33 & 81.47 & 70.99
\end{tabular}
\caption{\label{tbl-res-nn}
Comparison in generalization performance of benchmark NLP systems with
a vanilla neural network (NN) approach, on POS, chunking, NER and SRL tasks. We
report results with both the word-level log-likelihood (WLL) 
and the sentence-level log-likelihood (SLL).
Generalization performance is reported in per-word accuracy
rate (PWA) for POS and F1 score for other tasks.
 The NN results are behind the benchmark results,
in \voirsec{sec-lm} we show how to improve these models using unlabeled data.}
\end{center}
\end{table}

\begin{table}[tb]
\renewcommand\arraystretch{1.2}
\begin{center}
\begin{tabular}{cccccc}
Task & Window/Conv. size & Word dim. & Caps dim. & Hidden units & Learning rate \\ \hline
POS & $d_{win} = 5$ & $d^0 = 50$ &  $d^1=5$ & $n_{hu}^1 = 300$ & $\lambda = 0.01$ \\ \hdashline[.3pt/2pt]
CHUNK & " & " & " & " & " \\ \hdashline[.3pt/2pt]
NER & " & " & " & " & " \\ \hdashline[.3pt/2pt]
\multirow{2}{*}{SRL} & \multirow{2}{*}{"} & \multirow{2}{*}{"} & \multirow{2}{*}{"} & \multirow{2}{*}{$\begin{array}{c}n_{hu}^1 = 300\\ n_{hu}^2 = 500\end{array}$} & \multirow{2}{*}{"} \\
& & & & &
\end{tabular}
\caption{\label{tbl-hyper-parameters}
Hyper-parameters of our networks. We report for each task the window size
(or convolution size), word feature dimension, capital feature dimension,
number of hidden units and learning rate.}
\renewcommand\arraystretch{1}
\end{center}
\end{table}

For POS, chunking and NER tasks, we report results with the window
architecture described in~\voirsec{sec-window-approach}. The SRL task was
trained using the sentence approach (\voirsec{sec-sentence-approach}).
Results are reported in~\voirtbl{tbl-res-nn}, in per-word accuracy (PWA) for
POS, and F1 score for all the other tasks. We performed experiments both
with the word-level log-likelihood (WLL) and with the sentence-level log-likelihood (SLL).
The hyper-parameters of our networks are reported
in~\voirtbl{tbl-hyper-parameters}.  All our networks were fed with two raw
text features: lower case words, and a capital letter feature. We chose to
consider lower case words to limit the number of words in the
dictionary. However, to keep some upper case information lost by this
transformation, we added a ``caps'' feature which tells if each word was in
low caps, was all caps, had first letter capital, or had one capital.
Additionally, all occurrences of sequences of numbers within a word are replaced with the string ``NUMBER'', so for example both the words ``PS1'' and ``PS2'' would map to the single word ``psNUMBER''.
We used a dictionary containing the 100,000 most common words in WSJ (case insensitive).
Words outside this dictionary were replaced by a single special ``RARE'' word.

Results show that neural networks ``out-of-the-box'' are behind baseline
benchmark systems. Looking at all submitted systems reported on each CoNLL
challenge website showed us our networks performance are nevertheless in
the performance ballpark of existing approaches.
The training criterion which takes into account the sentence structure (SLL)
seems to boost the performance for the Chunking, NER and SRL
tasks, with little advantage for POS. This result is in line with existing
NLP studies comparing sentence-level and word-level likelihoods~\citep{liang:2008}.
The capacity of our network architectures lies mainly in the word lookup table,
which contains $50\times 100,000$ parameters to train. In the WSJ data, $15\%$
of the most common words appear about $90\%$ of the time. Many words appear
only a few times. It is thus very difficult to train properly their corresponding
50 dimensional feature vectors in the lookup table. Ideally, we would like
semantically similar words to be close in the embedding space represented
by the word lookup table: by continuity of the neural network function,
tags produced on semantically similar sentences would be similar. We show
in~\voirtbl{tbl-embeddings-nolm} that it is not the case: neighboring words
in the embedding space do not seem to be semantically related.

\begin{table}[tb]
\small
\begin{center}
\begin{sc}
\begin{tabular}{cccccc}
france     &  jesus       & xbox     & reddish     & scratched   & megabits     \\
454        &  1973        & 6909     & 11724       & 29869       & 87025        \\ \hline
persuade   & thickets     & decadent & widescreen  & odd         & ppa          \\
faw        & savary       & divo     & antica      & anchieta    & uddin        \\
blackstock & sympathetic  & verus    & shabby      & emigration  & biologically \\
giorgi     & jfk          & oxide    & awe         & marking     & kayak        \\
shaheed    & khwarazm     & urbina   & thud        & heuer       & mclarens     \\
rumelia    & stationery   & epos     & occupant    & sambhaji    & gladwin      \\
planum     & ilias        & eglinton & revised     & worshippers & centrally    \\
goa'uld    & gsNUMBER     & edging   & leavened    & ritsuko     & indonesia    \\
collation  & operator     & frg      & pandionidae & lifeless    & moneo        \\
bacha      & w.j.         & namsos   & shirt       & mahan       & nilgiris    
\end{tabular}
\end{sc}
\caption{\label{tbl-embeddings-nolm}
Word embeddings in the word lookup table of a SRL neural network trained
from scratch, with a dictionary of size $100,000$.  For each column the
queried word is followed by its index in the dictionary (higher means more
rare) and its $10$ nearest neighbors (arbitrary using the Euclidean
metric).}
\end{center}
\bigskip
\end{table}

We will focus in the next section on 
improving these word
embeddings by leveraging unlabeled data. 
We will see our approach results 
in a performance boost for all tasks.

\begin{figure}
\subfloat[POS]{\includegraphics[width=0.24\linewidth]{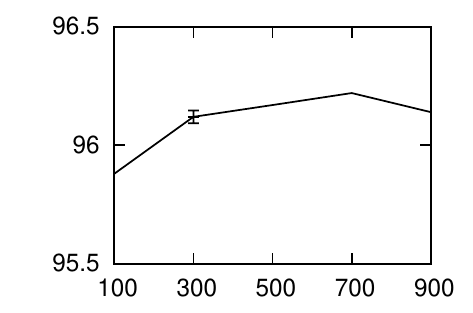}}
\subfloat[CHUNK]{\includegraphics[width=0.24\linewidth]{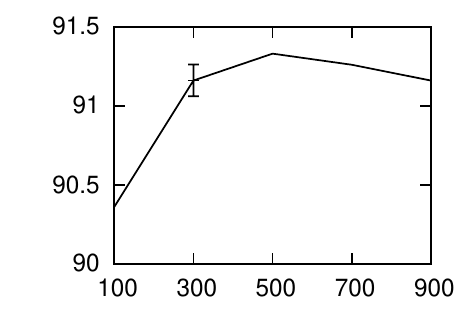}}
\subfloat[NER]{\includegraphics[width=0.24\linewidth]{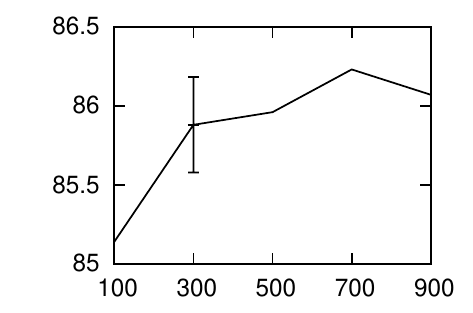}}
\subfloat[SRL]{\includegraphics[width=0.24\linewidth]{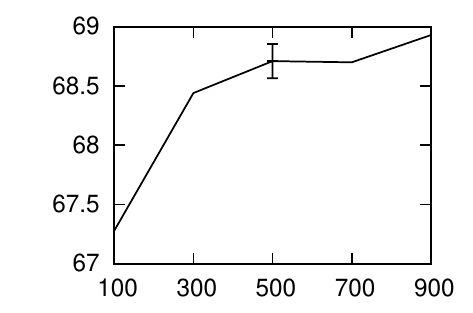}}
\caption{\label{fig-f1-vs-nhu} F1 score on the \emph{validation} set
  (y-axis) versus number of hidden units (x-axis) for different tasks
  trained with the sentence-level likelihood (SLL), as
  in~\voirtbl{tbl-res-nn}. For SRL, we vary in this graph only the number
  of hidden units in the second layer. The scale is adapted for each
  task. We show the standard deviation (obtained over 5 runs with different
  random initialization), for the architecture we picked (300 hidden units
  for POS, CHUNK and NER, 500 for SRL).  }
\end{figure}

\begin{remark}[Architectures]
In all our experiments in this paper, we tuned the hyper-parameters by
trying only a few different architectures by validation.
In practice, the choice of hyperparameters such as the number of hidden
units, provided they are large enough, has a limited impact on the
generalization performance.
In~\voirfig{fig-f1-vs-nhu}, we report the F1 score for each task on the
validation set, with respect to the number of hidden units. Considering the
variance related to the network initialization, we chose the smallest
network achieving ``reasonable'' performance, rather than picking the
network achieving the top performance obtained on a single run.
\end{remark}

\begin{remark}[Training Time]
Training our network is quite computationally expensive. Chunking and NER
take about one hour to train, POS takes few hours, and SRL takes about
three days. Training could be faster with a larger learning rate, but we
prefered to stick to a small one which works, rather than finding the
optimal one for speed.  Second order methods~\citep{lecun:1998} could be
another speedup technique.
\end{remark}
\bigskip


\section{Lots of Unlabeled Data}
\label{sec-lm}

We would like to obtain word embeddings
carrying more syntactic and semantic information
than shown in~\voirtbl{tbl-embeddings-nolm}.
Since most of the trainable parameters of our system
are associated with the word embeddings,
these poor results suggest that we should use
considerably more training data.
Following our NLP \emph{from scratch} philosophy, 
we now describe how to dramatically improve these 
embeddings using large unlabeled datasets.
We then use these improved embeddings to initialize
the word lookup tables of the networks 
described in \voirsec{sec-supervised-results}.

\subsection{Datasets}

Our first English corpus is the entire English
Wikipedia.\footnote{Available at \url{http://download.wikimedia.org}. We
  took the November 2007 version.}
We have removed all paragraphs containing
non-roman characters and all MediaWiki markups. The resulting text was
tokenized using the Penn Treebank tokenizer script.\footnote{Available
at \url{http://www.cis.upenn.edu/~treebank/tokenization.html}.}  
The resulting dataset contains about $631$ million words.  
As in our previous experiments, we use a dictionary containing 
the 100,000 most common words in WSJ, with the same processing of capitals and numbers. 
Again, words outside the dictionary 
were replaced by the special ``RARE'' word. 

Our second English corpus is composed by adding 
an extra $221$ million words extracted from 
the Reuters~RCV1~\citep{lewis:2004} dataset.\footnote{Now available 
at \url{http://trec.nist.gov/data/reuters/reuters.html}.}
We also extended the dictionary to $130,000$ words by
adding the $30,000$ most common words in Reuters.
This is useful in order to determine whether improvements
can be achieved by further increasing the unlabeled dataset size.

\subsection{Ranking Criterion versus Entropy Criterion}
\label{sec-ranking}

We used these unlabeled datasets to train  \emph{language models} 
that compute \emph{scores} describing the acceptability 
of a piece of text.  These language models are again
large neural networks using the window approach
described in~\voirsec{sec-window-approach} and
in~\voirfig{fig-net-window}.  As in the previous section,
most of the trainable parameters are 
located in the lookup tables. 

Similar language models were already proposed by \citet{yoshua:2001}
and \citet{schwenk:2002}.  Their goal was to estimate the \emph{probability}
of a word given the previous words in a sentence. Estimating conditional
probabilities suggests a cross-entropy criterion similar to those described
in~\voirsec{sec-isolated-tag-criterion}.  Because the dictionary size is
large, computing the normalization term can be extremely demanding, and
sophisticated approximations are required.  More importantly for us, neither
work leads to significant word embeddings being reported.

\citet{shannon:1951} has estimated the entropy of the English language 
between 0.6 and 1.3 bits per character by asking human subjects to guess upcoming
characters. \citet{cover-king:1978} give a lower bound of 1.25 bits per
character using a subtle gambling approach.  Meanwhile, using a simple word trigram
model, \citet{brown-1992} reach 1.75 bits per character.
\citet{teahan-1996} obtain entropies as low as 1.46 bits per character
using variable length character $n$-grams.  The human subjects rely of course
on all their knowledge of the language and of the world.  Can we learn the
grammatical structure of the English language and the nature of the world by
leveraging the 0.2 bits per character that separate human subjects from simple
n-gram models?  Since such tasks certainly require high capacity models,
obtaining sufficiently small confidence intervals on the test set entropy may
require prohibitively large training sets.\footnote{However, \citet{klein-manning:2001}
describe a rare example of realistic unsupervised grammar induction
using a cross-entropy approach on binary-branching parsing trees,
that is, by forcing the system to generate a hierarchical representation.}
The entropy criterion lacks dynamical range because its numerical value is largely
determined by the most frequent phrases. In order to learn syntax, 
rare but legal phrases are no less significant than common phrases.

It is therefore desirable to define alternative training criteria.
We propose here to use a \emph{pairwise ranking} approach \citep{cohen-1998}.  
We seek a network that computes a higher score when given a legal phrase than when
given an incorrect phrase.  Because the ranking literature often deals with
information retrieval applications, many authors define complex ranking
criteria that give more weight to the ordering of the best ranking
instances \citep[see][]{burges-2007,clemencon-2007}.  
However, in our case, we do not want to emphasize the most common phrase
over the rare but legal phrases. Therefore we use a simple pairwise criterion.  

We consider a \emph{window} approach network, as described
in~\voirsec{sec-window-approach} and~\voirfig{fig-net-window}, with
parameters ${\bm \theta}$ which outputs a score $\f{}(\bm{x})$
given a window of text ${\bm x} = \seq{w}{1}{d_{win}}$. We minimize
the ranking criterion with respect to $\bm{\theta}$:
\begin{equation}
\small
\label{eq-lm-cost}
 \bm{\theta} \mapsto \sum_{\bm{x} \in {\cal X}} \sum_{w\in {\cal D}} 
    \max\left\{\:0\:,\: 1 - \f{}(\bm{x}) + \f{}(\bm{x}^{(w)}) \:\right\}\,,
\end{equation}
where ${\cal X}$ is the set of all possible text windows with $d_{win}$ words coming from our training corpus,
${\cal D}$ is the dictionary of words, and $\bm{x}^{(w)}$ denotes the text window
obtained by replacing the central word of text window $\seq{w}{1}{d_{win}}$ by the word $w$.

\citet{okanohara:2007} use a related approach to avoiding the entropy
criteria using a binary classification approach (correct/incorrect
phrase). Their work focuses on using a kernel classifier, and not on
learning word embeddings as we do here.
\citet{smith:2005} also propose a contrastive criterion which estimates the
likelihood of the data conditioned to a ``negative'' neighborhood. They
consider various data neighborhoods, including sentences of length $d_{win}$
drawn from ${\cal D}^{d_{win}}$. Their goal was however to  perform well on 
some tagging task on fully unsupervised data, rather than obtaining generic
word embeddings useful for other tasks.

\subsection{Training Language Models}
\label{sec-training-language-models}

The language model network was trained by stochastic gradient minimization of
the ranking criterion~\voireq{eq-lm-cost}, sampling a sentence-word pair
$(s,\,w)$ at each iteration.

\medskip

Since training times for such large scale systems are counted in weeks, it is
not feasible to try many combinations of hyperparameters.  It also makes sense
to speed up the training time by initializing new networks with the embeddings
computed by earlier networks. In particular, we found it expedient to train a
succession of networks using increasingly large dictionaries, each network
being initialized with the embeddings of the previous network. Successive
dictionary sizes and switching times are chosen
arbitrarily. \citep{bengio:2009} provides a more detailed discussion of this,
the (as yet, poorly understood) ``curriculum'' process.

For the purposes of model selection we use the process of ``breeding''.
The idea of breeding is instead of trying a full grid search of possible values (which we did
not have enough computing power for) to search for the parameters in anology to breeding
biological cell lines. 
Within each line, child networks are initialized with the
embeddings of their parents and trained on increasingly rich datasets 
with sometimes different parameters.
That is, suppose we have $k$ processors, which is much less than the possible set of 
parameters one would like to try.
One chooses $k$ initial parameter choices from the large set, and trains these on the 
$k$ processors.  In our case, possible parameters to adjust are:
 the  learning rate $\lambda$, the word embedding dimensions $d$, 
number of hidden units $n_{hu}^1$ and input window size $d_{win}$.
 One then trains each of these models in an online
fashion for a certain amount of time (i.e. a few days), and then selects the best ones using
the validation set error rate. 
That is, breeding decisions were made on 
the basis of the value of the ranking criterion~\voireq{eq-lm-cost} 
estimated on a validation set composed of one million words held out 
from the Wikipedia corpus.
In the next breeding iteration, one then chooses another set of $k$ parameters from the 
possible grid of values that permute slightly the most successful candidates from the previous 
round.  As many of these parameter choices can share weights, we can effectively continue online
training retaining some of the learning from the previous iterations.

Very long training times make such strategies necessary for the foreseeable
future: if we had been given computers ten times faster, we probably would
have found uses for datasets ten times bigger. 
However, we should say we believe that although we ended up with a particular choice of parameters,
many other choices are almost equally as good, although perhaps there are others that are better
as we could not do a full grid search. 


\medskip

In the following subsections, 
we report results obtained with two trained language models. 
The results achieved by these two models are representative
of those achieved by networks trained on the full corpuses.
\begin{itemize}
\item
   Language model LM1 has a window size $d_{win} = 11$ 
   and a hidden layer with $n_{hu}^1 = 100$ units.
   The embedding layers were dimensioned like those
   of the supervised networks (\voirtbl{tbl-hyper-parameters}).
   Model LM1 was trained on our first English corpus (Wikipedia) using successive 
   dictionaries composed of the $5000$, $10,000$, $30,000$, $50,000$ 
   and finally $100,000$ most common WSJ words. 
   The total training time was about four weeks.
\item
   Language model LM2 has the same dimensions.
   It was initialized with the embeddings of LM1,
   and trained for an additional three weeks on 
   our second English corpus (Wikipedia+Reuters) using a 
   dictionary size of 130,000 words.
\end{itemize}

\subsection{Embeddings}

\begin{table}[tb]
\small
\begin{center}
\begin{sc}
\begin{tabular}{cccccc}
france      & jesus   & xbox        & reddish   & scratched & megabits   \\
454         & 1973    & 6909        & 11724     & 29869     & 87025      \\ \hline
austria     & god     & amiga       & greenish  & nailed    & octets     \\
belgium     & sati    & playstation & bluish    & smashed   & mb/s       \\
germany     & christ  & msx         & pinkish   & punched   & bit/s      \\
italy       & satan   & ipod        & purplish  & popped    & baud       \\
greece      & kali    & sega        & brownish  & crimped   & carats     \\
sweden      & indra   & psNUMBER    & greyish   & scraped   & kbit/s     \\
norway      & vishnu  & hd          & grayish   & screwed   & megahertz  \\
europe      & ananda  & dreamcast   & whitish   & sectioned & megapixels \\
hungary     & parvati & geforce     & silvery   & slashed   & gbit/s     \\
switzerland & grace   & capcom      & yellowish & ripped    & amperes    
\end{tabular}
\end{sc}
\caption{\label{tbl-embeddings-lm}
Word embeddings in the word lookup table of the language model neural
network LM1 trained with a dictionary of size $100,000$. For each column the
queried word is followed by its index in the dictionary (higher means more
rare) and its $10$ nearest neighbors (using the Euclidean
metric, which was chosen arbitrarily).}
\end{center}
\bigskip
\end{table}

Both networks produce much more appealing word embeddings than in Section~\ref{before-semi-sup}.
\voirtbl{tbl-embeddings-lm} shows the ten nearest neighbors 
of a few randomly chosen query words for the LM1 model.  The syntactic and
semantic properties of the neighbors are clearly related to those of the
query word.  These results are far more satisfactory than those reported
in~\voirtbl{tbl-embeddings-lm} for embeddings obtained using purely supervised
training of the benchmark NLP tasks.

\subsection{Semi-supervised Benchmark Results}
\label{sec-semi-supervised-results}

Semi-supervised learning has been the object of much attention 
during the last few years~\citep[see][]{chapelle:2006}.
Previous semi-supervised approaches for NLP 
can be roughly categorized as follows:
\begin{itemize}
\item
  Ad-hoc approaches such as \citep{rosenfeld:2007} 
  for relation extraction.
\item
  Self-training approaches,
  such as~\citep{ueffing:2007} for machine translation,
  and~\citep{mcclosky:2008} for parsing.
  These methods augment the labeled training set 
  with examples from the unlabeled dataset using 
  the labels predicted by the model itself.
  Transductive approaches, such as~\citep{joachims:1999a} 
  for text classification can be viewed as a refined
  form of self-training.
\item
  Parameter sharing approaches such as \citep{ando:2005,suzuki:2008}.
  \citeauthor{ando:2005} propose a multi-task approach where they jointly
  train models sharing certain parameters.  They train POS and NER models
  together with a language model (trained on 15 million words)
  consisting of predicting words given the surrounding
  tokens. \citeauthor{suzuki:2008} embed a generative model (Hidden Markov
  Model) inside a CRF for POS, Chunking and NER. The generative model is
  trained on one billion words.
  These approaches should be seen as a linear counterpart of our work.
  Using multilayer models vastly expands the parameter sharing 
  opportunities (see~\voirsec{sec-multi-task}).
\end{itemize}

Our approach simply consists of initializing the word lookup 
tables of the supervised networks with the embeddings computed by 
the language models. Supervised training is then performed 
as in  \voirsec{sec-supervised-results}.
In particular the supervised training stage is free
to modify the lookup tables. This sequential approach is 
computationally convenient because it separates the lengthy 
training of the language models from the relatively fast 
training of the supervised networks.  Once the language
models are trained, we can perform multiple experiments
on the supervised networks in a relatively short time.
Note that our procedure is clearly linked to the (semi-supervised) 
deep learning procedures of \citep{hinton,bengio-deep,weston2008deep}.

\begin{table}
\begin{center}
\begin{tabular}{l|c|c|c|c}
{\bf Approach}    & {\bf POS}  & {\bf CHUNK} & {\bf NER} & {\bf SRL} \\
            & (PWA) & (F1) & (F1) & (F1) \\ \hline
\textbf{Benchmark Systems} & {97.24} & {94.29} & {89.31} & {77.92} \\ \hline 
NN+WLL          & 96.31 & 89.13 & 79.53 & 55.40 \\
NN+SLL          & 96.37  & 90.33 & 81.47 & 70.99 \\
\hline
NN+WLL+LM1      & 97.05 & 91.91 & 85.68 & 58.18 \\
NN+SLL+LM1      & 97.10 & 93.65 & 87.58 & 73.84 \\
\hline
NN+WLL+LM2      & 97.14 & 92.04 & 86.96 & 58.34 \\
NN+SLL+LM2      & 97.20 & 93.63 & 88.67 & 74.15
\end{tabular}
\caption{\label{tbl-res-nn-lm}
Comparison in generalization performance of benchmark  NLP systems with
our (NN) approach on POS, chunking, NER and SRL tasks. We
report results with both the word-level log-likelihood (WLL) and the sentence-level log-likelihood (SLL).
We report with (LM$n$) performance of the networks
trained from the language model embeddings (\voirtbl{tbl-embeddings-lm}).
Generalization performance is reported in per-word accuracy (PWA) for POS and
F1 score for other tasks.}
\end{center}
\end{table}

\voirtbl{tbl-res-nn-lm} clearly shows that this 
simple initialization significantly boosts 
the generalization performance of the supervised
networks for each task. It is worth mentioning the larger 
language model led to even better performance.
This suggests that we could still take advantage of even bigger unlabeled datasets.

\subsection{Ranking and Language}
\label{sec-zelig}

There is a large agreement in the NLP community that 
syntax is a necessary prerequisite for semantic 
role labeling~\citep{gildea:2002}. This is why 
state-of-the-art semantic role labeling systems 
thoroughly exploit multiple parse trees.
The parsers themselves~\citep{charniak:2000,collins:1999}
contain considerable prior information about syntax (one can 
think of this as a kind of informed pre-processing).

Our system does not use such parse trees because 
we attempt to learn this information from the 
unlabeled data set. It is therefore legitimate to question
whether our ranking criterion~\voireq{eq-lm-cost} has the 
conceptual capability to capture such a rich 
hierarchical information. 
At first glance, the ranking task appears unrelated to the 
induction of probabilistic grammars that underly standard parsing 
algorithms. The lack of hierarchical representation seems a fatal
flaw~\citep{chomsky:1956}.  

However, ranking is closely related to
an alternative description of the language structure:
\emph{operator grammars}~\citep{harris:1968}.
Instead of directly studying the structure of a sentence, 
\citeauthor{harris:1968} defines an algebraic structure 
on the space of all sentences. Starting from a couple of elementary 
sentence forms, sentences are described by the successive 
application of sentence transformation operators. 
The sentence structure is revealed as a side effect of the 
successive transformations. Sentence transformations 
can also have a semantic interpretation.

In the spirit of structural linguistics,
\citeauthor{harris:1968} describes procedures to
discover sentence transformation operators by
leveraging the statistical regularities of the language.
Such procedures are obviously useful for machine learning approaches.
In particular, he proposes a test to decide whether two 
sentences forms are semantically related by a transformation operator.
He first defines a ranking criterion~\citep[section~4.1]{harris:1968}:
\begin{quotation}
  ``Starting for convenience with very short sentence forms, say $ABC$,
    we choose a particular word choice for all the classes, say $B_qC_q$,
    except one, in this case $A$; for every pair of members $A_i$, $A_j$
    of that word class we ask how the sentence formed with one of the
    members, \ie. $A_iB_qC_q$ compares as to acceptability
    with the sentence formed with the other member, 
    \ie. $A_jB_qC_q$.'' 
\end{quotation}
These \emph{gradings} are then used to compare sentence forms:
\begin{quotation}
  ``It now turns out that, given the graded $n$-tuples of words
    for a particular sentence form, we can find other sentences
    forms of the same word classes in which the same $n$-tuples
    of words produce the same grading of sentences.'' 
\end{quotation}
This is an indication that these two sentence forms
exploit common words with the same syntactic function
and possibly the same meaning.
This observation forms the empirical basis
for the construction of operator grammars that describe 
real-world natural languages such as English.

Therefore there are solid reasons to believe that the ranking
criterion~\voireq{eq-lm-cost} has the conceptual potential to capture strong
syntactic and semantic information. On the other hand, the structure of our 
language models is probably too restrictive for such goals,
and our current approach only exploits the
word embeddings discovered during training.

\section{Multi-Task Learning}
\label{sec-multi-task}

It is generally accepted that features \emph{trained} 
for one task can be useful for \emph{related tasks}.
This idea was already exploited in the previous section
when certain language model features, namely the word embeddings, 
were used to initialize the supervised networks.

Multi-task learning (MTL) leverages this idea in a more systematic way.
Models for all tasks of interests are \emph{jointly trained} with 
an additional linkage between their trainable parameters
in the hope of improving the generalization error.
This linkage can take the form of a regularization term in the joint cost 
function that biases the models towards common representations.
A much simpler approach consists in having the models 
{\emph{share certain parameters}} defined a priori.
Multi-task learning has a long history in machine 
learning and neural networks.
\citet{caruana:1997} gives a good overview 
of these past efforts.

\subsection{Joint Decoding versus Joint Training}

Multitask approaches do not necessarily involve joint training. For instance,
modern speech recognition systems use Bayes rule to combine the outputs of an
acoustic model trained on speech data and a language model trained on phonetic
or textual corpora~\citep{jelinek-1976}.  This joint decoding approach has
been successfully applied to structurally more complex NLP tasks.
\citet{sutton:2005a} obtains improved results by combining the predictions 
of independently trained CRF models using a joint decoding process at 
test time that requires more sophisticated probabilistic inference techniques.
On the other hand, \citet{sutton:2005} obtain results somewhat below the
state-of-the-art using joint decoding for SRL and syntactic parsing.
\citet{musillo:2006} also describe a negative result at the same joint task.

Joint decoding invariably works by considering additional probabilistic
dependency paths between the models. Therefore it defines an implicit
supermodel that describes all the tasks in the same probabilistic
framework. Separately training a submodel only makes sense when the
training data blocks these additional dependency paths (in the sense of
d-separation,~\citealp{pearl:1988}).  This implies that, 
without joint training, the additional dependency paths 
cannot directly involve unobserved variables.
Therefore, the natural idea of discovering common internal 
representations across tasks requires joint training.

Joint training is relatively straightforward when the training sets for the
individual tasks contain the same patterns with different labels.  It is then
sufficient to train a model that computes multiple outputs for each
pattern~\citep{suddarth-1991}.  Using this scheme, \citet{sutton:2007}
demonstrates improvements on POS tagging and noun-phrase chunking using
jointly trained CRFs.  However the joint labeling requirement is a limitation
because such data is not often available. \citet{miller:2000} achieves
performance improvements by jointly training NER, parsing, and relation
extraction in a statistical parsing model.  The joint labeling requirement
problem was weakened using a predictor to fill in the missing annotations.

\citet{ando:2005} propose a setup that works 
around the joint labeling requirements.  
They define linear models of the form
\mbox{$f_i(x)=\bm{w}_i^\top\:\Phi(x)+\bm{v}_i^\top\:\Theta\Psi(x)$}
where $f_i$ is the classifier for the $i$-th task
with parameters $\bm{w}_i$ and $\bm{v}_i$.  
Notations $\Phi(x)$ and $\Psi(x)$ represent 
engineered features for the pattern $x$.
Matrix $\Theta$ maps the $\Psi(x)$ features 
into a low dimensional subspace common across all tasks.
Each task is trained using its own examples 
without a joint labeling requirement.
The learning procedure alternates
the optimization of $\bm{w}_i$ and $\bm{v}_i$ for each task, 
and the optimization of $\Theta$ to minimize the average
loss for all examples in all tasks.
The authors also consider auxiliary unsupervised tasks
for predicting substructures. 
They report excellent results on several tasks, 
including POS and NER.

\subsection{Multi-Task Benchmark Results}

\begin{figure}[tb]
\begin{center}
 \includegraphics{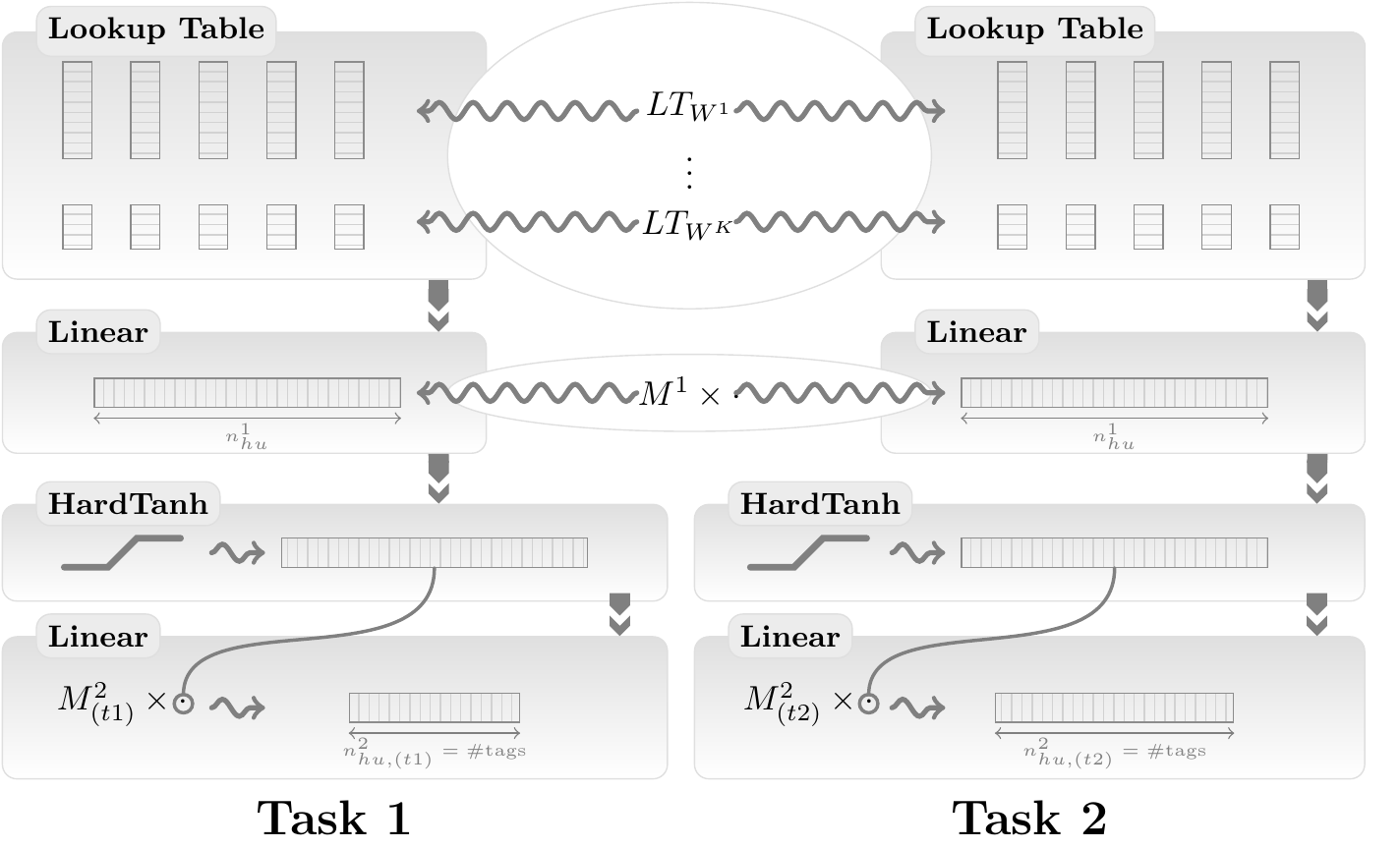}
 \caption{\label{fig-mtl}Example of multitasking with NN. 
    Task 1 and Task 2 are two tasks trained with the window approach architecture
    presented in Figure~\ref{fig-net-window}. 
    Lookup tables as well as the first hidden layer are shared. 
    The last layer is task specific. The principle is the
    same with more than two tasks.}
\end{center}
\end{figure}

\begin{table}
\begin{center}
\begin{tabular}{l|c|c|c|c}
{\bf Approach}    & {\bf POS}  & {\bf CHUNK} & {\bf NER} & {\bf SRL} \\
            & (PWA) & (F1) & (F1) & (F1) \\ \hline
\textbf{Benchmark Systems} & {97.24} & {94.29} & {89.31} & -- \\ \hline 
& \multicolumn{4}{c}{\em Window Approach} \\
NN+SLL+LM2           & 97.20 & 93.63 & 88.67 & -- \\
NN+SLL+LM2+MTL       & 97.22  & 94.10 & 88.62 & -- \\ \hline
& \multicolumn{4}{c}{\em Sentence Approach} \\
NN+SLL+LM2           & 97.12 & 93.37 & 88.78 & 74.15 \\
NN+SLL+LM2+MTL       & 97.22  & 93.75 & 88.27 & 74.29
\end{tabular}
\caption{\label{tbl-res-nn-mtl} 
  Effect of multi-tasking on our neural architectures. We trained POS,
  CHUNK NER in a MTL way, both for the window and sentence network
  approaches. SRL was only included in the sentence approach joint
  training. As a baseline, we show previous results of our window approach
  system, as well as additional results for our sentence approach system,
  when trained separately on each task. Benchmark system performance is also
  given for comparison.
}
\end{center}
\bigskip
\end{table}

\voirtbl{tbl-res-nn-mtl} reports results obtained by jointly trained models
for the POS, CHUNK, NER and SRL tasks using the same setup
as~\voirsec{sec-semi-supervised-results}. We trained jointly POS, CHUNK and
NER using the window approach network. As we mentioned earlier, SRL can be
trained only with the sentence approach network, due to long-range
dependencies related to the verb predicate. We thus also trained all four
tasks using the sentence approach network. In both cases, all models share
the lookup table parameters~\voireq{eq-lookup-table-multi}.
The parameters of the first linear layers~\voireq{eq-linear-layer} were
shared in the window approach case (see~\voirfig{fig-mtl}), and the first
the convolution layer parameters~\voireq{eq-convolution-layer} were shared
in the sentence approach networks.

For the window approach, best results were obtained by enlarging the first
hidden layer size to $n_{hu}^1 = 500$ (chosen by validation) in order to
account for its shared responsibilities.  We used the same architecture
than SRL for the sentence approach network.
The word embedding dimension was kept constant $d^0=50$ in order to reuse
the language models of~\voirsec{sec-semi-supervised-results}.

Training was achieved by minimizing the 
loss averaged across all tasks.
This is easily achieved with stochastic gradient
by alternatively picking examples for each task
and applying~\voireq{eq-criterion-derivative}
to all the parameters of the corresponding model,
including the shared parameters. Note that this gives each task equal weight.
Since each task uses the training sets 
described in~\voirtbl{tbl-experimental-setup},
it is worth noticing that examples can come from quite 
different datasets. The generalization performance for each task
was measured using the traditional testing data specified
in~\voirtbl{tbl-experimental-setup}. Fortunately, 
none of the training and test sets overlap across tasks.

While we find worth mentioning that MTL can produce a single unified
architecture that performs well
for all these tasks, no (or only marginal) improvements were obtained with
this approach compared to training separate architectures per task (which still use
 semi-supervised learning, which is somehow the most important MTL task). 
The next section shows we can leverage known correlations
between tasks in more direct manner.




\section{The Temptation}
\label{sec-nlp-not-from-scratch}

Results so far have been obtained by staying 
(almost\footnote{We did some basic preprocessing 
of the raw input words as 
described in \voirsec{before-semi-sup}, hence the ``almost'' in the title of this article. A completely from scratch approach would presumably not know anything about words at all and would work from letters only (or, taken to a further extreme, from speech or optical character recognition, as humans do).})  true to our \emph{from scratch}
philosophy.  We have so far avoided specializing our architecture for any
task, disregarding a lot of useful \emph{a priori} NLP knowledge.  We have
shown that, thanks to large unlabeled datasets, our generic neural
networks can still achieve close to state-of-the-art 
performance by discovering useful features.
This section explores what happens when we increase the 
level of task-specific engineering in our systems by incorporating some 
common techniques from the NLP literature.
We often obtain further improvements.
These figures are useful to quantify how far
we went by leveraging large datasets instead
of relying on a priori knowledge.

\begin{table}[tb]
\begin{center}
\begin{tabular}{l|c|c|c|c}
{\bf Approach}    & {\bf POS}  & {\bf CHUNK} & {\bf NER} & {\bf SRL} \\
            & (PWA) & (F1) & (F1) \\ \hline
\textbf{Benchmark Systems} 
  & {97.24} & {94.29} & {89.31} & {77.92} \\
\hline 
NN+SLL+LM2           & 97.20 & 93.63 & 88.67 & 74.15 \\
\hline
NN+SLL+LM2+Suffix2   & 97.29 & --    & --    & -- \\
NN+SLL+LM2+Gazetteer & --    & --    & 89.59 & -- \\
NN+SLL+LM2+POS       & --    & 94.32 & 88.67 & -- \\
NN+SLL+LM2+CHUNK     & --    & --    & --    & 74.72 \\
\end{tabular}
\caption{\label{tbl-cascade-basic}
Comparison in generalization performance of 
benchmark NLP systems with our neural networks (NNs)
using increasing task-specific engineering.
We report results
obtained with a network trained without the extra task-specific features 
(\voirsec{sec-multi-task})
and with the extra task-specific features  described in \voirsec{sec-nlp-not-from-scratch}. 
The POS network was trained with two character word suffixes;
the NER network was trained using the small CoNLL 2003 gazetteer;
the CHUNK and NER networks were trained with additional POS features;
and finally, the SRL network was trained with additional CHUNK features.}
\end{center}
\end{table}

\subsection{Suffix Features}

Word suffixes in many western languages are strong 
predictors of the syntactic function of the word
and therefore can benefit the POS system.
For instance, \citet{ratnaparkhi:1996} uses inputs
representing word suffixes and prefixes up to four characters.
We achieve this in the POS task by adding discrete word 
features (\voirsec{sec-discrete-features}) representing the 
last two characters of every word. 
The size of the suffix dictionary was $455$.
This led to a small improvement of the POS performance
(\voirtbl{tbl-cascade-basic}, row {\small NN+SLL+LM2+Suffix2}).
We also tried suffixes obtained with the \citet{porter:1980} stemmer
and obtained the same performance as when using 
two character suffixes.

\subsection{Gazetteers}

State-of-the-art NER systems often use a large dictionary 
containing well known named entities~\citep[\eg.][]{florian:2003}.
We restricted ourselves to the gazetteer provided by the CoNLL challenge,
containing $8,000$ locations, person names, organizations, and miscellaneous
entities.  
We trained a NER network with 4 additional word features indicating (feature ``on'' or ``off'')
whether the word is found in the gazetteer under one of these four categories.
The gazetteer includes not only words, but also chunks of words. If a
sentence chunk is found in the gazetteer, then all words in the chunk
have their corresponding gazetteer feature turned to ``on''.
The resulting system displays a clear performance improvement
(\voirtbl{tbl-cascade-basic}, row {\small NN+SLL+LM2+Gazetteer}),
slightly outperforming the baseline.
A plausible explanation of this large boost over the network using only the
language model is that gazeetters include word chunks, while we use only
the word representation of our language model. For example, ``united'' and
``bicycle'' seen separately are likely to be non-entities, while ``united
bicycle'' might be an entity, but catching it would require higher level
representations of our language model.

\subsection{Cascading}

When one considers related tasks, 
it is reasonable to assume that tags obtained for one task can 
be useful for taking decisions in the other tasks. 
Conventional NLP systems often use features obtained 
from the output of other preexisting NLP systems.
For instance, \citet{shen:2005} describe a chunking
system that uses POS tags as input; 
\citet{florian:2003} describes a NER system whose 
inputs include POS and CHUNK tags,
as well as the output of two other NER classifiers.
State-of-the-art SRL systems exploit
parse trees \citep{gildea:2002,punyakanok:2005},
related to CHUNK tags, and built 
using POS tags~\citep{charniak:2000,collins:1999}.

\voirtbl{tbl-cascade-basic} reports results obtained
for the CHUNK and NER tasks by adding discrete 
word features (\voirsec{sec-discrete-features})
representing the POS tags. In order to facilitate comparisons,
instead of using the more accurate tags from 
our POS network, we use for each task the POS tags 
provided by the corresponding CoNLL challenge.
We also report results obtained for the SRL task by 
adding word features representing the CHUNK tags (also provided by the
CoNLL challenge).
We consistently obtain moderate improvements.


\subsection{Ensembles}

\begin{table}[tb]
\begin{center}
\begin{tabular}{ll|c|c|c}
{\bf Approach}   && {\bf POS} & {\bf CHUNK} & {\bf NER} \\
                 && (PWA) & (F1) & (F1) \\ \hline
\multicolumn{2}{l|}{\textbf{Benchmark Systems}} 
                          & 97.24 & 94.29 &  89.31 \\ \hline 
NN+SLL+LM2+POS & worst    & 97.29 & 93.99  &  89.35  \\
NN+SLL+LM2+POS & mean     & 97.31 & 94.17  &  89.65  \\
NN+SLL+LM2+POS & best     & 97.35 & 94.32  &  89.86  \\ \hline
NN+SLL+LM2+POS & voting ensemble &  97.37 & 94.34 & 89.70 \\
NN+SLL+LM2+POS & joined ensemble & 97.30 & 94.35 & 89.67
\end{tabular}
\caption{\label{tbl-ensemble}
Comparison in generalization performance for POS, CHUNK and NER tasks
of the networks obtained using by combining 
ten training runs with different initialization.}
\end{center}
\end{table}

Constructing ensembles of classifiers is a 
proven way to trade computational efficiency
for generalization performance~\citep{bell-2007}.
Therefore it is not surprising that many NLP
systems achieve state-of-the-art performance
by combining the outputs of multiple classifiers.
For instance, \citet{kudoh:2001}
use an ensemble of classifiers trained with 
different tagging conventions (see \voirsec{sec-tags}).
Winning a challenge is of course a legitimate objective.
Yet it is often difficult to figure out which 
ideas are most responsible for the state-of-the-art 
performance of a large ensemble.

Because neural networks are nonconvex, training runs with different initial
parameters usually give different solutions.  \voirtbl{tbl-ensemble}
reports results obtained for the CHUNK and NER task after ten training runs
with random initial parameters. Voting the ten network outputs 
on a per tag basis (``voting ensemble'') leads to a small improvement over the
average network performance.  We have also tried a more sophisticated
ensemble approach: the ten network output scores (before sentence-level
likelihood) were combined with an additional linear
layer~\voireq{eq-linear-layer} and then fed to a new sentence-level
likelihood~\voireq{eq-sentence-likelihood}.  The parameters of 
the combining layers were then trained on the existing training set, 
while keeping the ten networks fixed (``joined ensemble'').
This approach did not improve on simple voting.


These ensembles come of course at the expense
of a ten fold increase of the running time.
On the other hand, multiple training times could 
be improved using smart sampling 
strategies~\citep{neal-1996}.

We can also observe that the performance
variability among the ten networks is not very large. 
The local minima found by the training 
algorithm are usually good local minima,
thanks to the oversized parameter space
and to the noise induced by the stochastic
gradient procedure~\citep{lecun:1998}.
In order to reduce the variance in our experimental
results, we always use the same initial parameters
for networks trained on the same task (except
of course for the results reported in \voirtbl{tbl-ensemble}.)

\subsection{Parsing}

\citet{gildea:2002} offer several arguments 
suggesting that syntactic parsing is a necessary 
prerequisite for the SRL task.
The CoNLL 2005 SRL benchmark task provides
parse trees computed using \emph{both} 
the \citet{charniak:2000} and \citet{collins:1999} parsers.
State-of-the-art systems often exploit additional 
parse trees such as the $k$ top ranking parse trees
\citep{koomen:2005,haghighi:2005}.

In contrast our SRL networks so far do not use parse trees at all.
They rely instead on internal representations transferred
from a language model trained with an objective function that 
captures a lot of syntactic information (see \voirsec{sec-zelig}).
It is therefore legitimate to question whether this 
approach is an acceptable lightweight replacement 
for parse trees. 

\begin{figure}[tb]
  \begin{center}
  \begin{tabular}{lc}
    {\parbox{4em}{\sc level $0$}} & 
    \parbox{.8\linewidth}{\hfil\includegraphics[scale=.8]{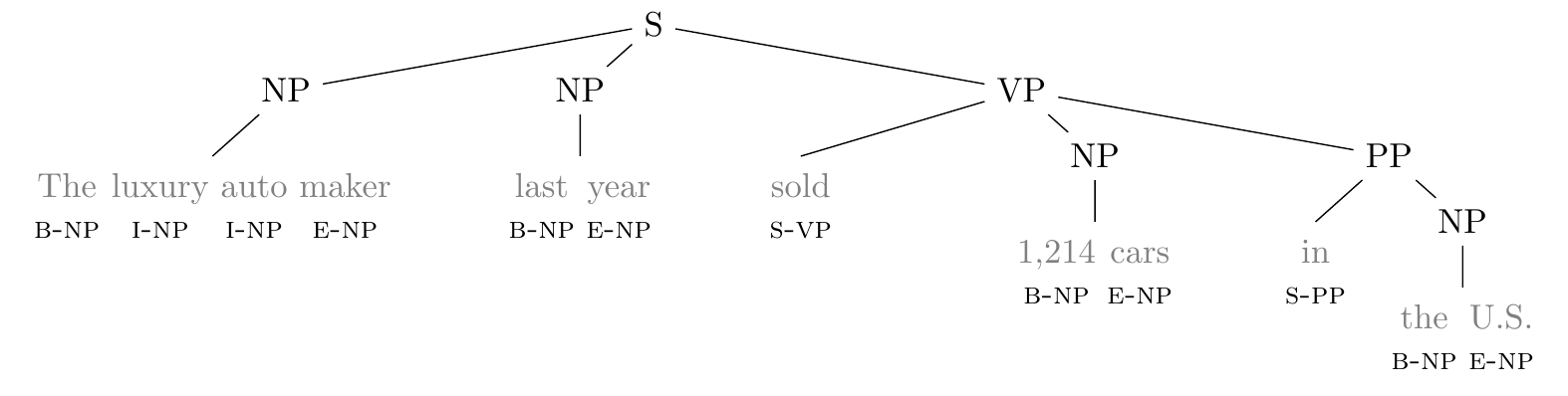}}\\
    {\parbox{4em}{\sc level $1$}} & 
    \parbox{.8\linewidth}{\hfil\includegraphics[scale=.8]{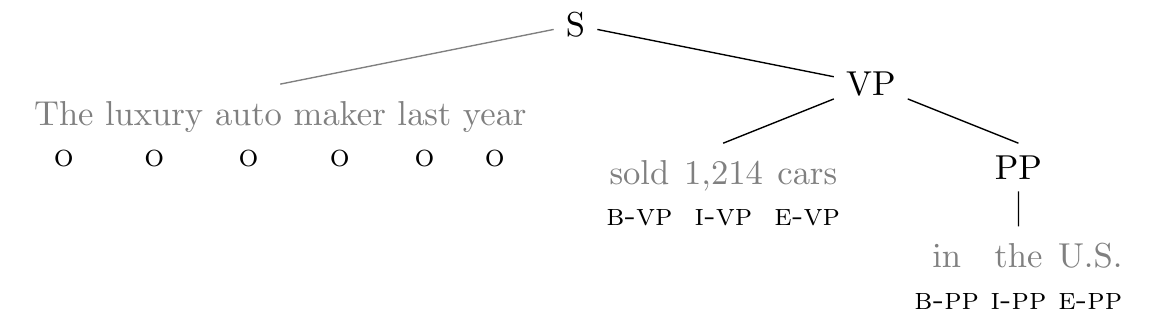}}\\
    {\parbox{4em}{\sc level $2$}} & 
    \parbox{.8\linewidth}{\hfil\includegraphics[scale=.8]{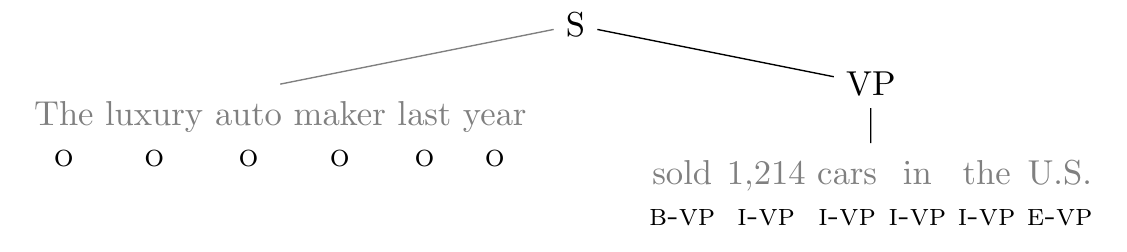}}
  \end{tabular}
  \caption{\label{fig-parse-tree-levels}
        Charniak parse tree for the sentence \emph{``The luxury auto maker
        last year sold 1,214 cars in the U.S.''}. Level~$0$ is the original
        tree. Levels~$1$ to~$4$ are obtained by successively collapsing 
        terminal tree branches. For each level, words receive tags
        describing the segment associated with the corresponding leaf.
        All words receive tag ``O'' at level 3 in this example.}
   \end{center}
   \bigskip
\end{figure}

We answer this question by providing parse tree information as additional
input features to our system. We have limited ourselves to the Charniak
parse tree provided with the CoNLL 2005 data. Considering that a node in a
syntactic parse tree assigns a label to a segment of the parsed sentence,
we propose a way to feed (partially) this labeled segmentation to our
network, through additional lookup tables. Each of these lookup tables
encode \emph{labeled} segments of each parse tree level (up to a certain
depth). The labeled segments are fed to the network following a IOBES
tagging scheme (see~Sections~\ref{sec-tags}
and~\ref{sec-discrete-features}).  As there are $40$ different phrase
labels in WSJ, each additional tree-related lookup tables has $161$ entries
($40\times 4 + 1$) corresponding to the IBES segment tags, plus the extra O
tag.

We call level $0$ the information associated with the leaves of the
original Charniak parse tree. The lookup table for level $0$ encodes the
corresponding IOBES phrase tags for each words. We obtain levels $1$ to $4$
by repeatedly trimming the leaves as shown
in~\voirfig{fig-parse-tree-levels}.  We labeled ``O'' words belonging to
the root node ``S'', or all words of the sentence if the root itself has
been trimmed.

\begin{table}[htb!]
\begin{center}
\iffalse
\begin{tabular}{l|c}
{\bf Approach} & {\bf SRL} \\ \hline
\textbf{Benchmark System} (six parse trees) & {77.92} \\
NN+SLL+LM2    & 74.15 \\ \hline
NN+SLL+LM2+Charniak (level $0$ only) & 75.65 \\
NN+SLL+LM2+Charniak (levels $0$ \& $1$) & 75.81 \\
NN+SLL+LM2+Charniak (levels $0$ to $2$) & 76.05 \\
NN+SLL+LM2+Charniak (levels $0$ to $3$) & 75.89 \\
NN+SLL+LM2+Charniak (levels $0$ to $4$) & 76.06
\end{tabular}
\else
\begin{tabular}{l|cc}
{\bf Approach} & \multicolumn{2}{c}{\bf SRL} \\ \hline
               & (valid) & (test) \\ \hline
\textbf{Benchmark System} (six parse trees)               & 77.35 & 77.92 \\
\textbf{Benchmark System} (top Charniak parse tree only)  & 74.76 & -- \\ \hline 
NN+SLL+LM2                                                & 72.29 & 74.15 \\ \hline
NN+SLL+LM2+Charniak (level $0$ only)                      & 74.44 & 75.65 \\
NN+SLL+LM2+Charniak (levels $0$ \& $1$)                   & 74.50 & 75.81 \\
NN+SLL+LM2+Charniak (levels $0$ to $2$)                   & 75.09 & 76.05 \\
NN+SLL+LM2+Charniak (levels $0$ to $3$)                   & 75.12 & 75.89 \\
NN+SLL+LM2+Charniak (levels $0$ to $4$)                   & 75.42 & 76.06 \\ \hline
NN+SLL+LM2+CHUNK                                          & --    & 74.72 \\
NN+SLL+LM2+PT0                                            & --    & 75.49
\end{tabular}
\fi
\caption{\label{tbl-res-nn-parse-srl} Generalization performance on the SRL
  task of our NN architecture compared with the benchmark system. We show
  performance of our system fed with different levels of depth of the
  Charniak parse tree. We report previous results of our architecture with
  no parse tree as a baseline.  \citet{koomen:2005} report test and
  validation performance using six parse trees, as well as validation
  performance using only the top Charniak parse tree. For comparison
  purposes, we hence also report validation performance. Finally, we report our
  performance with the CHUNK feature, and compare it against a level 0 feature PT0
  obtained by our network.
}
\end{center}
\bigskip
\end{table}

Experiments were performed using the LM2 language model using the same network
architectures (see~\voirtbl{tbl-hyper-parameters}) and using additional lookup
tables of dimension $5$ for each parse tree level.
\voirtbl{tbl-res-nn-parse-srl} reports the performance
improvements obtained by providing increasing levels
of parse tree information.
Level $0$ alone increases the F1 score by almost $1.5\%$.  Additional
levels yield diminishing returns.  The top performance reaches $76.06\%$ F1
score.  This is not too far from the state-of-the-art system which we note uses
six parse trees instead of one. \citet{koomen:2005} also report a
$74.76\%$ F1 score on the validation set using only the Charniak parse
tree.
Using the first three parse tree levels, we reach this performance on the validation set.

We also reported in~\voirtbl{tbl-res-nn-parse-srl} our previous performance
obtained with the CHUNK feature (see \voirtbl{tbl-cascade-basic}). It is
surprising to observe that adding chunking features into the semantic role
labeling network performs significantly worse than adding features
describing the level $0$ of the Charniak parse tree
(\voirtbl{tbl-res-nn-parse-srl}).
Indeed, if we ignore the label prefixes ``BIES'' defining the segmentation,
the parse tree leaves (at level $0$) and the chunking have identical
labeling. However, the parse trees identify leaf sentence segments that are
often smaller than those identified by the chunking tags, as shown
by~\citet{hollingshead:2005}.\footnote{As in~\citep{hollingshead:2005},
  consider the sentence and chunk labels ``(NP They) (VP are starting to
  buy) (NP growth stocks)''. The parse tree can be written as ``(S (NP
  They) (VP are (VP starting (S (VP to (VP buy (NP growth stocks)))))))''.
  The tree leaves segmentation is thus given by ``(NP They) (VP are) (VP
  starting) (VP to) (VP buy) (NP growth stocks)''.
}
Instead of relying on Charniak parser, we chose to train a second chunking
network to identify the segments delimited by the leaves of the Penn
Treebank parse trees (level $0$). Our network achieved $92.25\%$ F1 score
on this task (we call it PT0), while we evaluated Charniak performance as
$91.94\%$ on the same task. As shown in~\voirtbl{tbl-res-nn-parse-srl},
feeding our own PT0 prediction into the SRL system
 gives similar performance to using
Charniak predictions, and is consistently better than the CHUNK feature.

\subsection{Word Representations}

In~\voirsec{sec-lm}, we adapted our neural network architecture for
training a language model task. By leveraging a large amount of unlabeled
text data, we induced word embeddings which were shown to boost
generalization performance on all tasks. While we chose to stick with one
single architecture, other ways to induce word representations exist.
\citet{mnih:2007} proposed a related language model
approach inspired from Restricted Boltzmann Machines. 
However, 
word representations are perhaps more commonly infered from $n$-gram language
modelling rather than smoothed language models. One popular approach is the
Brown clustering algorithm~\citep{brown:1992}, which builds hierachical
word clusters by maximizing the bigram's mutual information. The induced word
representation has been used with success in a wide variety of NLP tasks,
including POS~\citep{schutze:1995}, NER~\citep{miller:2004,ratinov:2009}, or
parsing~\citep{koo:2008}. Other related approaches exist, like phrase
clustering~\citep{lin:2009} which has been shown to work well for NER. Finally,
\citet{huang:2009} have recently proposed a smoothed language modelling
approach based on a Hidden Markov Model, with success on POS and Chunking tasks.

While a comparison of all these word representations is beyond the scope of
this paper, it is rather fair to question the quality of our word
embeddings compared to a popular NLP approach. In this section, we report a
comparison of our word embeddings against Brown clusters, when used as
features into our neural network architecture. We report as baseline
previous results where our word embeddings are \emph{fine-tuned} for each
task. We also report performance when our embeddings are kept fixed during
task-specific training. Since \emph{convex} machine learning algorithms are
common practice in NLP, we finally report performances for the convex
version of our architecture.

For the convex experiments, we considered the linear version of our neural
networks (instead of having several linear layers interleaved with a
non-linearity). While we always picked the sentence approach for SRL, we
had to consider the window approach in this particular convex setup, as the
sentence approach network (see~\voirfig{fig-net-sentence}) includes a Max
layer. Having only one linear layer in our neural network is not enough to
make our architecture convex: all lookup-tables (for each discrete feature)
must also be \emph{fixed}. The word-lookup table is simply fixed to the
embeddings obtained from our language model LM2. All other discrete feature
lookup-tables (caps, POS, Brown Clusters...) are fixed to a standard
\emph{sparse} representation. Using the notation introduced
in~\voirsec{sec-discrete-features}, if $LT_{W^k}$ is the lookup-table of
the $k^{th}$ discrete feature, we have $W^k \in \R^{|{\cal D}^k|\times
  |{\cal D}^k|}$ and the representation of the discrete input $w$ is
obtained with:
\begin{equation}
\label{eq-sparse-lookup}
LT_{W^k}(w) = \col{W^k}{w}\, = \left( 0, \cdots 0, \, \underset{{\rm at\ index\ } w}{1},  \, 0, \, \cdots \, 0 \right)^{\T}\,.
\end{equation}
Training our architecture in this convex setup with the sentence-level
likelihood~\voireq{eq-sentence-likelihood} corresponds to training a
CRF. In that respect, these convex experiments show the performance of our
word embeddings in a classical NLP framework.

Following the \citet{ratinov:2009} and \citet{koo:2008} setups, we generated
$1,000$ Brown clusters using the implementation\footnote{Available
at~\url{http://www.eecs.berkeley.edu/~pliang/software}.}
from~\citet{liang:2005}.
To make the comparison fair, the clusters were first induced on the
concatenation of Wikipedia and Reuters datasets, as we did
in~\voirsec{sec-lm} for training our largest language model LM2, using a
130K word dictionary. This dictionary covers about $99\%$ of the 
words in the training set of each task. To cover the
last $1\%$, we augmented the dictionary with the missing words (reaching
about 140K words) and induced Brown Clusters using the concatenation of
WSJ, Wikipedia, and Reuters.

The Brown clustering approach is hierarchical and generates a binary tree
of clusters. Each word in the vocabulary is assigned to a node in the
tree. Features are extracted from this tree by considering the path from
the root to the node containing the word of interest.  Following Ratinov \&
Roth, we picked as features the path prefixes of size $4$, $6$, $10$ and
$20$. In the non-convex experiments, we fed these four Brown Cluster
features to our architecture using four different lookup tables, replacing
our word lookup table. The size of the lookup tables was chosen to be $12$
by validation. In the convex case, we used the classical sparse
representation~\voireq{eq-sparse-lookup}, as for any other discrete feature.

\begin{table}[tb]
\begin{center}
\begin{tabular}{l|c|c|c|c}
{\bf Approach}    & {\bf POS}  & {\bf CHUNK} & {\bf NER} & {\bf SRL} \\
            & (PWA) & (F1) & (F1) & (F1) \\ \hline 
& \multicolumn{4}{c}{\emph{Non-convex Approach}} \\
LM2 (non-linear NN)                        & 97.29 & 94.32 & 89.59 & 76.06 \\
LM2 (non-linear NN, fixed embeddings)      & 97.10 & 94.45 & 88.79 & 72.24 \\
Brown Clusters (non-linear NN, 130K words) & 96.92 & 94.35 & 87.15 & 72.09 \\
Brown Clusters (non-linear NN, all words)  & 96.81 & 94.21 & 86.68 & 71.44 \\ \hline
& \multicolumn{4}{c}{\emph{Convex Approach}} \\
LM2 (linear NN, fixed embeddings)         & 96.69 & 93.51 & 86.64 & 59.11 \\
Brown Clusters (linear NN, 130K words)    & 96.56 & 94.20 & 86.46 & 51.54 \\
Brown Clusters (linear NN, all words)     & 96.28 & 94.22 & 86.63 & 56.42
\end{tabular}
\caption{\label{tbl-brown-clusters} Generalization performance of our
  neural network architecture trained with our language model (LM2) word
  embeddings, and with the word representations derived from the Brown
  Clusters. As before, all networks are fed with a capitalization
  feature. Additionally, POS is using a word suffix of size 2 feature,
  CHUNK is fed with POS, NER uses the CoNLL 2003 gazetteer, and SRL is fed
  with levels 1--5 of the Charniak parse tree, as well as a verb position
  feature. We report performance with both convex and non-convex
  architectures (300 hidden units for all tasks, with an additional 500
  hidden units layer for SRL). We also provide results for Brown Clusters
  induced with a 130K word dictionary, as well as Brown Clusters induced
  with all words of the given tasks.  }
\end{center}
\end{table}

We first report in~\voirtbl{tbl-brown-clusters} generalization performance
of our best non-convex networks trained with our LM2 language model and
with Brown Cluster features. Our embeddings perform at least as well as
Brown Clusters. Results are more mitigated in a convex setup. For most
task, going non-convex is better for both word representation types.
In general, ``fine-tuning'' our embeddings for each task also gives an
extra boost.  Finally, using a better word coverage with Brown Clusters
(``all words'' instead of ``130K words'' in~\voirtbl{tbl-brown-clusters})
did not help.

More complex features could be possibly combined instead of using a
non-linear model.  For instance, \citet{turian:2010} performed a comparison of
Brown Clusters and embeddings trained in the same spirit as
ours\footnote{%
However they did not reach our embedding performance. There are
several differences in how they trained their models that might
explain this. Firstly, they may have experienced difficulties because they train
50-dimensional embeddings for 269K distinct words using a
comparatively small training set (RCV1, 37M words), unlikely to
contain enough instances of the rare words. Secondly, they predict the
correctness of the final word of each window instead of the center
word~\citep{turian:2010}, effectively restricting the model to
unidirectional prediction. Finally, they do not fine tune their
embeddings after unsupervised training.

},
with additional features combining labels and tokens. We believe this type
of comparison should be taken with care, as combining a given feature with
different word representations might not have the same effect on each word
representation.

\subsection{Engineering a Sweet Spot}
\label{sec-senna}

We implemented a standalone version of our architecture, written in the C
language. We gave the name ``SENNA'' (Semantic/syntactic Extraction using a
Neural Network Architecture) to the resulting system. The parameters of
each architecture are the ones described
in~\voirtbl{tbl-hyper-parameters}. All the networks were trained separately
on each task using the sentence-level likelihood (SLL). The word embeddings
were initialized to LM2 embeddings, and then fine-tuned for each task. We
summarize features used by our implementation
in~\voirtbl{tbl-senna-features}, and we report performance achieved on each
task in~\voirtbl{tbl-sweet-spot}.
\begin{table}
  \center
  \begin{tabular}{l|l}
    Task & Features \\ \hline
    POS   & Suffix of size 2 \\
    CHUNK & POS \\
    NER   & CoNLL 2003 gazetteer \\
    PT0   & POS \\
    SRL   & PT0, verb position \\
  \end{tabular}
\caption{\label{tbl-senna-features} Features used by SENNA implementation, for each task. In addition, all tasks use ``low caps word'' and ``caps'' features.
}
\end{table}
\begin{table}[tb]
  \begin{center}
  \begin{tabular}{ll|c|c}
     \bf Task & & \bf Benchmark & \bf SENNA \\
     \hline
        Part of Speech (POS)   & (Accuracy)   & 97.24 \% & 97.29 \% \\
        Chunking (CHUNK)               & (F1) & 94.29 \% & 94.32 \% \\
        Named Entity Recognition (NER) & (F1) & 89.31 \% & 89.59 \% \\
        Parse Tree level 0 (PT0)       & (F1) & 91.94 \% & 92.25 \% \\
        Semantic Role Labeling (SRL)  & (F1) & 77.92 \% & 75.49 \%
  \end{tabular}
  \caption{\label{tbl-sweet-spot}
  Performance of the engineered sweet spot (SENNA) on various tagging tasks.
  The PT0 task replicates the sentence segmentation of the parse tree leaves.  
  The corresponding benchmark score measures the quality of the 
  Charniak parse tree leaves relative to the Penn Treebank gold parse trees.}
  \end{center}
\end{table}
\begin{table}[tb]
\centering
\begin{tabular}{c|rr}
\bf POS System & \bf RAM (MB) & \bf Time (s) \\ \hline
\citet{toutanova:2003} & 800 & 64 \\
\citet{shen:2007} & 2200 & 833 \\ \hline
SENNA & 32 & 4\\
\multicolumn{3}{c}{}\\
\bf SRL System & \bf RAM (MB) & \bf Time (s) \\ \hline
\citet{koomen:2005} & 3400 & 6253 \\ \hline
SENNA & 124 & 51
\end{tabular}
\caption{\label{tbl-pos-runtime} Runtime speed and memory consumption
  comparison between state-of-the-art systems and our approach (SENNA). We
  give the runtime in seconds for running both the POS and SRL taggers on
  their respective testing sets. Memory usage is reported in megabytes.}
\end{table}
The runtime version\footnote{Available at \url{http://ml.nec-labs.com/senna}.}
contains about 2500 lines of C code, runs in less than 150MB of memory,
and needs less than a millisecond per word to compute all the tags.
\voirtbl{tbl-pos-runtime} compares the tagging speeds for our system
and for the few available state-of-the-art systems:
the~\citet{toutanova:2003} POS tagger\footnote{Available
at \url{http://nlp.stanford.edu/software/tagger.shtml}. We picked the 3.0 version (May 2010).}, the~\citet{shen:2007} POS
tagger\footnote{Available at \url{http://www.cis.upenn.edu/~xtag/spinal}.}
and the~\citet{koomen:2005} SRL system.\footnote{Available
at \url{http://l2r.cs.uiuc.edu/~cogcomp/asoftware.php?skey=SRL}.}
All programs were run on a single 3GHz Intel core. 
The POS taggers were run with Sun Java 1.6 with a large enough
memory allocation to reach their top tagging speed.  
The beam size of the Shen tagger was set to $3$ 
as recommended in the paper.
Regardless of implementation differences, 
it is clear that our neural networks run considerably faster.  
They also require much less memory.
Our POS and SRL tagger runs in 32MB and 120MB of RAM respectively.
The Shen and Toutanova taggers slow down
significantly when the Java machine is given less 
than 2.2GB and 800MB of RAM respectively, while the Koomen tagger
requires at least 3GB of RAM.

We believe that a number of reasons 
explain the speed advantage of our system.
First, our system only uses rather simple input features
and therefore avoids the nonnegligible computation time
associated with complex handcrafted features.
Secondly, most network computations are \emph{dense} 
matrix-vector operations. In contrast, systems that rely
on a great number of \emph{sparse} features experience 
memory latencies when traversing the sparse data structures. 
Finally, our compact implementation is self-contained. 
Since it does not rely on the outputs
of disparate NLP system, it does not suffer from 
communication latency issues.


\section{Critical Discussion}

Although we believe that this contribution 
represents a step towards the ``NLP from scratch'' objective, 
we are keenly aware that both our goal and 
our means can be criticized. 

The main criticism of our goal can be summarized as follows.
Over the years, the NLP community has developed a considerable 
expertise in engineering effective NLP features.  
Why should they forget this painfully acquired expertise
and instead painfully acquire the skills required to train
large neural networks?  As mentioned in our introduction,
we observe that no single NLP task really covers the goals of NLP.
Therefore we believe that task-specific engineering 
(i.e. that does not generalize to other tasks) 
is not desirable.  But we also recognize how much our neural 
networks owe to previous NLP task-specific research.

The main criticism of our means is easier to address.
Why did we choose to rely on a twenty year
old technology, namely multilayer neural networks?
We were simply attracted by their ability to discover
hidden representations using a stochastic learning algorithm 
that scales linearly with the number of examples.
Most of the neural network technology necessary for our work
has been described ten years ago~\mbox{\citep[\eg.][]{lecun:1998b}}.
However, if we had decided ten years ago to train the language model 
network LM2 using a vintage computer, training would only be 
nearing completion today. Training algorithms that scale linearly 
are most able to benefit from such tremendous progress in
computer hardware.


\section{Conclusion}

We have presented a multilayer neural network architecture that
can handle a number of NLP tasks with both speed and accuracy.
The design of this system was determined by our desire
to avoid task-specific engineering as much as possible.
Instead we rely on large unlabeled datasets and 
let the training algorithm discover internal representations 
that prove useful for all the tasks of interest.
Using this strong basis, we have engineered
a fast and efficient ``all purpose'' NLP tagger
that we hope will prove useful to the community.

\bigskip\mbox{~}


\acks{
      We acknowledge the persistent support of NEC for this research effort.
      We thank Yoshua Bengio, Samy Bengio, Eric Cosatto, Vincent Etter,
      Hans-Peter Graf, Ralph Grishman, and Vladimir Vapnik
      for their useful feedback and comments.}

\newpage
\appendix
\section{Neural Network Gradients}
\label{apx-gradients}

We consider a neural network $f_{\bm{\theta}}(\cdot)$, with parameters
$\bm{\theta}$.  We maximize the likelihood~\voireq{eq-log-likelihood}, or
minimize ranking criterion~\voireq{eq-lm-cost}, with respect to the
parameters $\bm{\theta}$, using stochastic gradient. By negating the
likelihood, we now assume it all corresponds to minimize a cost
$C(f_{\bm{\theta}}(\cdot))$, with respect to $\bm{\theta}$.

Following the classical ``back-propagation''
derivations~\citep{lecun:1985,rumelhart:1986} and the modular approach
shown in~\citep{bottou:1991}, any feed-forward neural network with $L$
layers, like the ones shown in~\voirfig{fig-net-window}
and~\voirfig{fig-net-sentence}, can be seen as a composition of functions
$f^l_{\bm{\theta}}(\cdot)$, corresponding to each layer~$l$:
\begin{equation*}
  f_{\bm{\theta}}(\cdot) = f^L_{\bm{\theta}}(f^{L-1}_{\bm{\theta}}(\ldots f^1_{\bm{\theta}}(\cdot) \ldots))
\end{equation*}
Partionning the parameters of the network with respect to each layers $1 \leq l \leq L$, we write:
\begin{equation*}
\bm{\theta} = ( {\bm{\theta}}^1,\, \ldots, \,  \bm{\theta}^l,\, \ldots,\, \bm{\theta}^L )\,.
\end{equation*}
We are now interested in computing the gradients of the cost with respect
to each $\bm{\theta}^l$. Applying the chain rule (generalized to vectors) we obtain
the classical backpropagation recursion:
\begin{eqnarray}
\label{eq-backprop-params}
\frac{\partial C}{\partial \bm{\theta}^l} & = &
\frac{\partial f^l_{\bm{\theta}}}{\partial \bm{\theta}^l}
\,\frac{\partial C}{\partial f^l_{\bm{\theta}}} \\
\label{eq-backprop-inputs}
\frac{\partial C}{\partial f^{l-1}_{\bm{\theta}}} & = &
\frac{\partial f^l_{\bm{\theta}}}{\partial f^{l-1}_{\bm{\theta}}}
\,\frac{\partial C}{\partial f^{l}_{\bm{\theta}}}\,.
\end{eqnarray}
In other words, we first initialize the recursion by computing the
gradient of the cost with respect to the last layer output ${\partial
  C}/{\partial f^{L}_{\bm{\theta}}}$. Then each layer $l$ computes the
gradient respect to its own parameters with~\voireq{eq-backprop-params},
given the gradient coming from its output ${\partial C}/{\partial
  f^{l}_{\bm{\theta}}}$. To perform the backpropagation, it also computes
the gradient with respect to its own inputs, as shown
in~\voireq{eq-backprop-inputs}. We now derive the gradients for each layer
we used in this paper.
\paragraph{Lookup Table Layer} Given a matrix of parameters
$\bm{\theta}^1 = W^1$ and word (or discrete feature) indices
$\seq{w}{1}{T}$, the layer outputs the matrix:
\begin{equation*}
f^l_{\bm{\theta}}(\seq{w}{l}{T}) = 
\left( \begin{array}{cccc} \col{W}{\seq{w}{1}{}} & \col{W}{\seq{w}{2}{}} & \ldots & \col{W}{\seq{w}{T}{}} \end{array}\right)\,.
\end{equation*}
The gradients of the weights $\col{W}{i}$ are given by:
\begin{equation*}
\frac{\partial C}{\partial \col{W}{i}} = \sum_{\{1 \leq t \leq T\,/\,\seq{w}{t}{} = i\}} \window{\frac{\partial C}{\partial f^{l}_{\bm{\theta}}}}{i}{1}
\end{equation*}
This sum equals zero if the index $i$ in the lookup table does not
corresponds to a word in the sequence. In this case, the $i^{\textrm{th}}$ column
of $W$ does not need to be updated. As a Lookup Table Layer is always the
first layer, we do not need to compute its gradients with respect to the
inputs.
\paragraph{Linear Layer} Given parameters $\bm{\theta}^l = (W^l, \bm{b}^l)$, and an input \emph{vector} $f^{l-1}_{\bm{\theta}}$ the output is given by:
\begin{equation}
\label{eq-linear-forward}
f^l_{\bm{\theta}} = W^l f^{l-1}_{\bm{\theta}} + \bm{b}^l\,.
\end{equation}
The gradients with respect to the parameters are then obtained with:
\begin{equation}
\label{eq-linear-grad-params}
\frac{\partial C}{\partial W^l} = \left[ \frac{\partial C}{\partial f^l_{\bm{\theta}}} \right] \left[ f^{l-1}_{\bm{\theta}} \right]^{\T} \ \ \textrm{and} \ \
\frac{\partial C}{\partial \bm{b}^l} = \frac{\partial C}{\partial f^l_{\bm{\theta}}}\,,
\end{equation}
and the gradients with respect to the inputs are computed with:
\begin{equation}
\label{eq-linear-grad-inputs}
\frac{\partial C}{\partial f^{l-1}_{\bm{\theta}}} = \left[ W^l \right]^{\T} \frac{\partial C}{\partial f^l_{\bm{\theta}}}\,.
\end{equation}

\paragraph{Convolution Layer}
Given a input \emph{matrix} $f^{l-1}_{\bm{\theta}}$, a Convolution Layer
$f^l_{\bm{\theta}}(\cdot)$ applies a Linear Layer
operation~\voireq{eq-linear-forward} successively on each window
$\window{f^{l-1}_{\bm{\theta}}}{t}{d_{win}}\ (1 \leq t \leq T)$ of size
$d_{win}$. Using~\voireq{eq-linear-grad-params}, the gradients of the
parameters are thus given by summing over all windows:
$$\frac{\partial C}{\partial W^l} = \sum_{t=1}^T \left[ \window{\frac{\partial C}{\partial f^l_{\bm{\theta}}}}{t}{1} \right] \left[ \window{f^{l-1}_{\bm{\theta}}}{t}{d_{win}} \right]^{\T} \ \ \textrm{and} \ \
\frac{\partial C}{\partial \bm{b}^l} =  \sum_{t=1}^T \window{\frac{\partial C}{\partial f^l_{\bm{\theta}}}}{t}{1}\,.$$
After initializing the input gradients ${\partial C}/{\partial f^{l-1}_{\bm{\theta}}}$ to zero, we iterate~\voireq{eq-linear-grad-inputs}
over all windows for $1 \leq t \leq T$, leading the \emph{accumulation}\footnote{We denote ``$\plusequal$'' any accumulation operation.}:
$$\window{\frac{\partial C}{\partial f^{l-1}_{\bm{\theta}}}}{t}{d_{win}} \plusequal 
\left[ W^l \right]^{\T} \window{\frac{\partial C}{\partial f^l_{\bm{\theta}}}}{t}{1}\,.$$

\paragraph{Max Layer}
Given a \emph{matrix} $f^{l-1}_{\bm{\theta}}$, the Max Layer computes
$$
\left[ f^{l}_{\bm{\theta}} \right]_i = \max_{t} \left[ \window{f^{l-1}_{\bm{\theta}}}{t}{1} \right]_i \ \textrm{and} \ a_i = \argmax_{t} \left[ \window{f^{l-1}_{\bm{\theta}}}{t}{1} \right]_i \ \forall i \,,
$$
where $a_i$ stores the index of the largest value. We only need to compute the gradient with respect to the inputs, as this layer has no parameters. The gradient is given by
$$
\left[ \window{\frac{\partial C}{\partial f^{l-1}_{\bm{\theta}}}}{t}{1} \right]_i = \left\{ \begin{array}{cl} \left[ \window{\frac{\partial C}{ \partial f^{l}_{\bm{\theta}}}}{t}{1} \right]_i & \textrm{if} \ t = a_i \\ 0 & \textrm{otherwise} \end{array} \right. \,.
$$

\paragraph{HardTanh Layer}
Given a \emph{vector} $f^{l-1}_{\bm{\theta}}$, and the definition of the HardTanh~\voireq{eq-hardtanh-layer} we get
$$
 \left[ \frac{\partial C}{\partial f^{l-1}_{\bm{\theta}}} \right]_i = \left\{
\begin{array}{cl}
  0 & \textrm{if} \ \left[ f^{l-1}_{\bm{\theta}} \right]_i < -1 \\
  \left[ \frac{\partial C}{\partial f^{l}_{\bm{\theta}}} \right]_i & \textrm{if} \ -1 <= \left[ f^{l-1}_{\bm{\theta}} \right]_i <= 1 \\
  0 & \textrm{if} \ \left[ f^{l-1}_{\bm{\theta}} \right]_i > 1
\end{array}\right.\,,
$$
if we ignore non-differentiability points.

\paragraph{Word-Level Log-Likelihood}
The network outputs a score $\left[ f_{\bm{\theta}} \right]_i$ for each
tag indexed by $i$. Following~\voireq{eq-word-likelihood}, if $y$ is the true
tag for a given example, the stochastic score to minimize can be written as
$$
C(f_{\bm{\theta}}) = \logadd_j \left[ f_{\bm{\theta}} \right]_j - \left[ f_{\bm{\theta}} \right]_y
$$
Considering the definition of the $\logadd$~\voireq{eq-logadd}, the gradient with respect to $f_{\bm{\theta}}$ is given by
$$
\frac{\partial C}{\partial \left[ f_{\bm{\theta}} \right]_i } =
\frac{e^{\left[ f_{\bm{\theta}} \right]_i}}{\sum_k e^{\left[ f_{\bm{\theta}} \right]_k}} - \bm{1}_{i=y}\,\quad \forall i.
$$

\paragraph{Sentence-Level Log-Likelihood}
The network outputs a matrix where each element $\midx{\f{}}{i}{t}$ gives a score for tag $i$ at word $t$.
Given a tag sequence $\seq{y}{1}{T}$ and a input sequence $\seq{\bm{x}}{1}{T}$, we maximize the likelihood~\voireq{eq-sentence-likelihood},
which corresponds to minimizing the score
$$
C(f_{\bm{\theta}}, A) = 
\underbrace{\logadd_{\forall \seq{j}{1}{T}} s(\seq{\bm{x}}{1}{T},\, \seq{j}{1}{T},\, \tilde{\bm{\theta}})}_{C_{logadd}} \, - s(\seq{\bm{x}}{1}{T},\, \seq{y}{1}{T},\, \tilde{\bm{\theta}})\,,
$$
with
\begin{equation*}
s(\seq{\bm{x}}{1}{T},\, \seq{y}{1}{T},\, \tilde{\bm{\theta}}) = \sum_{t=1}^T \left( \midx{A}{\seq{y}{t-1}{}}{\seq{y}{t}{}} + \midx{\f{}}{\seq{y}{t}{}}{t}\right)\,.
\end{equation*}
We first initialize all gradients to zero
\begin{equation*}
\frac{\partial C}{\partial \midx{\f{}}{i}{t}} = 0 \ \forall{i,t}  \ \ \textrm{and} \ \ \frac{\partial C}{\partial \midx{A}{i}{j}} = 0 \quad \forall{i,j}\,.
\end{equation*}
We then \emph{accumulate} gradients over the second part of the cost $- s(\seq{\bm{x}}{1}{T},\, \seq{y}{1}{T},\, \tilde{\bm{\theta}})$, which gives:
\begin{equation*}
\begin{aligned}
\frac{\partial C}{\partial \midx{\f{}}{\seq{y}{t}{}}{t}} \, & \plusequal \,  1 \\
\frac{\partial C}{\partial \midx{A}{\seq{y}{t-1}{}}{\seq{y}{t}{}}} \,  & \plusequal \,  1
\end{aligned} \quad\quad \forall t\,.
\end{equation*}
We now need to accumulate the gradients over the first part of the cost, that is
$C_{logadd}$. We differentiate $C_{logadd}$ by applying the chain rule through the
recursion~\voireq{eq-forward-score-recursion}. First we initialize our recursion with
\begin{equation*}
\frac{\partial C_{logadd}}{\partial \delta_T(i)} = \frac{e^{\delta_T(i)}}{\sum_k e^{\delta_T(k)}} \quad \forall i\,. 
\end{equation*}
We then compute iteratively:
\begin{equation}
\frac{\partial C_{logadd}}{\partial \delta_{t-1}(i)} =  \sum_j \frac{\partial C_{logadd}}{\partial \delta_{t}(j)} \frac{e^{\delta_{t-1}(i) + \midx{A}{i}{j}}}{\sum_k e^{\delta_{t-1}(k) + \midx{A}{k}{j}}}\,,
\end{equation}
where at each step $t$ of the recursion we accumulate of the gradients with respect to the inputs $f_{\bm{\theta}}$, and the transition scores $\midx{A}{i}{j}$:
\begin{align*}
\frac{\partial C}{\partial \midx{\f{}}{i}{t}} & \plusequal \frac{\partial C_{logadd}}{\partial \delta_{t}(i)}\,\frac{\partial \delta_{t}(i)}{\partial \midx{\f{}}{i}{t}} & =  \frac{\partial C_{logadd}}{\partial \delta_{t}(i)} \\
\frac{\partial C}{\partial \midx{A}{i}{j}} & \plusequal  \frac{\partial C_{logadd}}{\partial \delta_{t}(j)}\,\frac{\partial \delta_{t}(j)}{\partial \midx{A}{i}{j}}
& =  \frac{\partial C_{logadd}}{\partial \delta_{t}(j)}\,\frac{e^{\delta_{t-1}(i) + \midx{A}{i}{j}}}{\sum_k e^{\delta_{t-1}(k) + \midx{A}{k}{j}}}\,.
\end{align*}

\paragraph{Ranking Criterion} We use the ranking criterion~\voireq{eq-lm-cost} for training our language model.
In this case, given a ``positive'' example $\bm{x}$ and a ``negative''
example $\bm{x^{(w)}}$, we want to minimize:
\begin{equation}
\label{eq-lm-cost-sto}
C(f_{\bm{\theta}}(\bm{x}), f_{\bm{\theta}}(\bm{x}^{w})) =
\max\left\{\:0\:,\: 1 - f_{\bm{\theta}}(\bm{x}) + f_{\bm{\theta}}(\bm{x}^{(w)}) \:\right\}\,.
\end{equation}
Ignoring the non-differentiability of $max(0,\cdot)$ in zero, the gradient is simply given by:
\begin{equation*}
\left(
\begin{array}{c}
\frac{\partial C}{\partial f_{\bm{\theta}}(\bm{x})} \\[0.2cm]
\frac{\partial C}{\partial f_{\bm{\theta}}(\bm{x}^w)}
\end{array}
\right) = 
\left\{
\begin{array}{cl}
  \left(\begin{array}{c} \scriptstyle -1 \\ \scriptstyle 1 \end{array}\right) & \textrm{if} \ 1 - f_{\bm{\theta}}(\bm{x}) + f_{\bm{\theta}}(\bm{x}^{(w)}) > 0\\
  \left(\begin{array}{c} \scriptstyle 0 \\ \scriptstyle 0 \end{array}\right) & \textrm{otherwise}
\end{array}
\right.\,.
\end{equation*}


\vskip 0.2in
\bibliography{all}

\begin{thebibliography}{89}
\providecommand{\natexlab}[1]{#1}
\providecommand{\url}[1]{\texttt{#1}}
\expandafter\ifx\csname urlstyle\endcsname\relax
  \providecommand{\doi}[1]{doi: #1}\else
  \providecommand{\doi}{doi: \begingroup \urlstyle{rm}\Url}\fi

\bibitem[Ando and Zhang(2005)]{ando:2005}
R.~K. Ando and T.~Zhang.
\newblock A framework for learning predictive structures from multiple tasks
  and unlabeled data.
\newblock \emph{JMLR}, 6:\penalty0 1817--1953, 11 2005.

\bibitem[Bell et~al.(2007)Bell, Koren, and Volinsky]{bell-2007}
R.~M. Bell, Y.~Koren, and C.~Volinsky.
\newblock The {BellKor} solution to the {Netflix Prize}.
\newblock Technical report, AT\&T Labs, 2007.
\newblock {\small\url{http://www.research.att.com/~volinsky/netflix}}.

\bibitem[Bengio and Ducharme(2001)]{yoshua:2001}
Y.~Bengio and R.~Ducharme.
\newblock A neural probabilistic language model.
\newblock In \emph{{NIPS} 13}, 2001.

\bibitem[Bengio et~al.(2007)Bengio, Lamblin, Popovici, and
  Larochelle]{bengio-deep}
Y.~Bengio, P.~Lamblin, D.~Popovici, and H.~Larochelle.
\newblock {Greedy layer-wise training of deep networks}.
\newblock In \emph{Advances in Neural Information Processing Systems, {NIPS}
  19}, 2007.

\bibitem[Bengio et~al.(2009)Bengio, Louradour, Collobert, and
  Weston]{bengio:2009}
Y.~Bengio, J.~Louradour, R.~Collobert, and J.~Weston.
\newblock Curriculum learning.
\newblock In \emph{International Conference on Machine Learning, {ICML}}, 2009.

\bibitem[Bottou(1991)]{bottou:1991}
L.~Bottou.
\newblock Stochastic gradient learning in neural networks.
\newblock In \emph{Proceedings of Neuro-N\^imes 91}, Nimes, France, 1991. EC2.

\bibitem[Bottou(1998)]{bottou-1998}
L.~Bottou.
\newblock Online algorithms and stochastic approximations.
\newblock In David Saad, editor, \emph{Online Learning and Neural Networks}.
  Cambridge University Press, Cambridge, UK, 1998.

\bibitem[Bottou and Gallinari(1991)]{bottou:1990}
L.~Bottou and P.~Gallinari.
\newblock A framework for the cooperation of learning algorithms.
\newblock In D.~Touretzky and R.~Lippmann, editors, \emph{Advances in Neural
  Information Processing Systems}, volume~3. Morgan Kaufmann, Denver, 1991.

\bibitem[Bottou et~al.(1997)Bottou, {LeCun}, and Bengio]{bottou-1997}
L.~Bottou, Y.~{LeCun}, and Yoshua Bengio.
\newblock Global training of document processing systems using graph
  transformer networks.
\newblock In \emph{Proc. of Computer Vision and Pattern Recognition}, pages
  489--493, Puerto-Rico, 1997. IEEE.

\bibitem[Bridle(1990)]{bridle:1990}
J.~S. Bridle.
\newblock Probabilistic interpretation of feedforward classification network
  outputs, with relationships to statistical pattern recognition.
\newblock In F.~Fogelman Souli\'e and J.~H\'erault, editors,
  \emph{Neurocomputing: Algorithms, Architectures and Applications}, pages
  227--236. NATO ASI Series, 1990.

\bibitem[Brown et~al.(1992{\natexlab{a}})Brown, deSouza, Mercer, Pietra, and
  Lai]{brown:1992}
P.~F. Brown, P.~V. deSouza, R.~L. Mercer, V.~J.~D. Pietra, and J~C. Lai.
\newblock Class-based n-gram models of natural language.
\newblock \emph{Computational Linguistics}, 18\penalty0 (4):\penalty0 467--479,
  1992{\natexlab{a}}.

\bibitem[Brown et~al.(1992{\natexlab{b}})Brown, Pietra, Mercer, Pietra, and
  Lai]{brown-1992}
P.~F. Brown, V.~J.~Della Pietra, R.~L. Mercer, S.~A.~Della Pietra, and J.~C.
  Lai.
\newblock An estimate of an upper bound for the entropy of english.
\newblock \emph{Computational Linguistics}, 18\penalty0 (1):\penalty0 31--41,
  1992{\natexlab{b}}.

\bibitem[Burges et~al.(2007)Burges, Ragno, and Le]{burges-2007}
C.~J.~C. Burges, R.~Ragno, and Quoc~Viet Le.
\newblock Learning to rank with nonsmooth cost functions.
\newblock In B.~Sch\"{o}lkopf, J.~Platt, and T.~Hoffman, editors,
  \emph{Advances in Neural Information Processing Systems 19}, pages 193--200.
  MIT Press, Cambridge, MA, 2007.

\bibitem[Caruana(1997)]{caruana:1997}
R.~Caruana.
\newblock {Multitask Learning}.
\newblock \emph{Machine Learning}, 28\penalty0 (1):\penalty0 41--75, 1997.

\bibitem[Chapelle et~al.(2006)Chapelle, Schölkopf, and Zien]{chapelle:2006}
O.~Chapelle, B.~Schölkopf, and A.~Zien.
\newblock \emph{Semi-Supervised Learning}.
\newblock Adaptive computation and machine learning. MIT Press, Cambridge,
  Mass., USA, 09 2006.

\bibitem[Charniak(2000)]{charniak:2000}
E.~Charniak.
\newblock {A maximum-entropy-inspired parser}.
\newblock \emph{Proceedings of the first conference on North American chapter
  of the Association for Computational Linguistics}, pages 132--139, 2000.

\bibitem[Chieu(2003)]{chieu:2003}
H.~L. Chieu.
\newblock Named entity recognition with a maximum entropy approach.
\newblock In \emph{In Proceedings of the Seventh Conference on Natural Language
  Learning (CoNLL-2003}, pages 160--163, 2003.

\bibitem[Chomsky(1956)]{chomsky:1956}
N.~Chomsky.
\newblock Three models for the description of language.
\newblock \emph{IRE Transactions on Information Theory}, 2\penalty0
  (3):\penalty0 113--124, September 1956.

\bibitem[Cl\'{e}men\c{c}on and Vayatis(2007)]{clemencon-2007}
S.~Cl\'{e}men\c{c}on and N.~Vayatis.
\newblock Ranking the best instances.
\newblock \emph{Journal of Machine Learning Research}, 8:\penalty0 2671--2699,
  December 2007.

\bibitem[Cohen et~al.(1998)Cohen, Schapire, and Singer]{cohen-1998}
W.~W. Cohen, R.~E. Schapire, and Y.~Singer.
\newblock Learning to order things.
\newblock \emph{Journal of Artificial Intelligence Research}, 10:\penalty0
  243--270, 1998.

\bibitem[Cohn and Blunsom(2005)]{cohn:2005}
T.~Cohn and P.~Blunsom.
\newblock Semantic role labelling with tree conditional random fields.
\newblock In \emph{Ninth Conference on Computational Natural Language (CoNLL)},
  2005.

\bibitem[Collins(1999)]{collins:1999}
M.~Collins.
\newblock \emph{Head-Driven Statistical Models for Natural Language Parsing}.
\newblock PhD thesis, University of Pennsylvania, 1999.

\bibitem[Collobert(2004)]{collobert:2004}
R.~Collobert.
\newblock \emph{Large Scale Machine Learning}.
\newblock PhD thesis, Universit\'e Paris {VI}, 2004.

\bibitem[Cover and King(1978)]{cover-king:1978}
T.~Cover and R.~King.
\newblock A convergent gambling estimate of the entropy of english.
\newblock \emph{IEEE Transactions on Information Theory}, 24\penalty0
  (4):\penalty0 413--421, July 1978.

\bibitem[Florian et~al.(2003)Florian, Ittycheriah, Jing, and
  Zhang]{florian:2003}
R.~Florian, A.~Ittycheriah, H.~Jing, and T.~Zhang.
\newblock Named entity recognition through classifier combination.
\newblock In \emph{Proceedings of the seventh conference on Natural language
  learning at HLT-NAACL 2003}, pages 168--171. Association for Computational
  Linguistics, 2003.

\bibitem[Gildea and Jurafsky(2002)]{gildea}
D.~Gildea and D.~Jurafsky.
\newblock {Automatic labeling of semantic roles}.
\newblock \emph{Computational Linguistics}, 28\penalty0 (3):\penalty0 245--288,
  2002.

\bibitem[Gildea and Palmer(2002)]{gildea:2002}
D.~Gildea and M.~Palmer.
\newblock The necessity of parsing for predicate argument recognition.
\newblock \emph{Proceedings of the 40th Annual Meeting of the ACL}, pages
  239--246, 2002.

\bibitem[Gim\'enez and M\`arquez(2004)]{gimenez:2004}
J.~Gim\'enez and L.~M\`arquez.
\newblock {SVMTool}: A general {POS} tagger generator based on support vector
  machines.
\newblock In \emph{Proceedings of the 4th International Conference on Language
  Resources and Evaluation ({LREC}'04)}, 2004.

\bibitem[Haghighi et~al.(2005)Haghighi, Toutanova, and Manning]{haghighi:2005}
A.~Haghighi, K.~Toutanova, and C.~D. Manning.
\newblock A joint model for semantic role labeling.
\newblock In \emph{Proceedings of the Ninth Conference on Computational Natural
  Language Learning (CoNLL-2005)}. Association for Computational Linguistics,
  June 2005.

\bibitem[Harris(1968)]{harris:1968}
Z.~S. Harris.
\newblock \emph{Mathematical Structures of Language}.
\newblock John Wiley \& Sons Inc., 1968.

\bibitem[Heckerman et~al.(2001)Heckerman, Chickering, Meek, Rounthwaite, and
  Kadie]{heckerman:2001}
D.~Heckerman, D.~M. Chickering, C.~Meek, R.~Rounthwaite, and C.~Kadie.
\newblock Dependency networks for inference, collaborative filtering, and data
  visualization.
\newblock \emph{Journal of Machine Learning Research}, 1:\penalty0 49--75,
  2001.
\newblock ISSN 1532-4435.

\bibitem[Hinton et~al.(2006)Hinton, Osindero, and Teh]{hinton}
G.~E. Hinton, S.~Osindero, and Y.-W. Teh.
\newblock A fast learning algorithm for deep belief nets.
\newblock \emph{Neural Comp.}, 18\penalty0 (7):\penalty0 1527--1554, July 2006.

\bibitem[Hollingshead et~al.(2005)Hollingshead, Fisher, and
  Roark]{hollingshead:2005}
K.~Hollingshead, S.~Fisher, and B.~Roark.
\newblock Comparing and combining finite-state and context-free parsers.
\newblock In \emph{HLT '05: Proceedings of the conference on Human Language
  Technology and Empirical Methods in Natural Language Processing}, pages
  787--794. Association for Computational Linguistics, 2005.

\bibitem[Huang and Yates(2009)]{huang:2009}
F.~Huang and A.~Yates.
\newblock Distributional representations for handling sparsity in supervised
  sequence-labeling.
\newblock In \emph{Proceedings of the Association for Computational Linguistics
  (ACL)}, pages 495--503. Association for Computational Linguistics, 2009.

\bibitem[Jelinek(1976)]{jelinek-1976}
F.~Jelinek.
\newblock Continuous speech recognition by statistical methods.
\newblock \emph{Proceedings of the IEEE}, 64\penalty0 (4):\penalty0 532--556,
  1976.

\bibitem[Joachims(1999)]{joachims:1999a}
T.~Joachims.
\newblock Transductive inference for text classification using support vector
  machines.
\newblock In \emph{{ICML}}, 1999.

\bibitem[Klein and Manning(2002)]{klein-manning:2001}
D.~Klein and C.~D. Manning.
\newblock Natural language grammar induction using a constituent-context model.
\newblock In Thomas~G. Dietterich, Suzanna Becker, and Zoubin Ghahramani,
  editors, \emph{Advances in Neural Information Processing Systems 14}, pages
  35--42. MIT Press, Cambridge, MA, 2002.

\bibitem[Koo et~al.(2008)Koo, Carreras, and Collins]{koo:2008}
T.~Koo, X.~Carreras, and M.~Collins.
\newblock Simple semi-supervised dependency parsing.
\newblock In \emph{Proceedings of the Association for Computational Linguistics
  (ACL)}, pages 595--603, 2008.

\bibitem[Koomen et~al.(2005)Koomen, Punyakanok, Roth, and Yih]{koomen:2005}
P.~Koomen, V.~Punyakanok, D.~Roth, and W.~Yih.
\newblock Generalized inference with multiple semantic role labeling systems
  (shared task paper).
\newblock In Ido Dagan and Dan Gildea, editors, \emph{Proc. of the Annual
  Conference on Computational Natural Language Learning (CoNLL)}, pages
  181--184, 2005.

\bibitem[Kudo and Matsumoto(2001)]{kudoh:2001}
T.~Kudo and Y.~Matsumoto.
\newblock Chunking with support vector machines.
\newblock In \emph{In Proceedings of the 2nd Meeting of the North American
  Association for Computational Linguistics: NAACL 2001}, pages 1--8.
  Association for Computational Linguistics, 2001.

\bibitem[Kudoh and Matsumoto(2000)]{kudoh:2000}
T.~Kudoh and Y.~Matsumoto.
\newblock Use of support vector learning for chunk identification.
\newblock In \emph{Proceedings of CoNLL-2000 and LLL-2000}, pages 142--144,
  2000.

\bibitem[Lafferty et~al.(2001)Lafferty, {McCallum}, and Pereira]{lafferty:2001}
J.~Lafferty, A.~{McCallum}, and F.~Pereira.
\newblock Conditional random fields: Probabilistic models for segmenting and
  labeling sequence data.
\newblock In \emph{Eighteenth International Conference on Machine Learning,
  {ICML}}, 2001.

\bibitem[{Le Cun} et~al.(1998){Le Cun}, Bottou, Bengio, and
  Haffner]{lecun:1998b}
Y.~{Le Cun}, L.~Bottou, Y.~Bengio, and P.~Haffner.
\newblock Gradient based learning applied to document recognition.
\newblock \emph{Proceedings of IEEE}, 86\penalty0 (11):\penalty0 2278--2324,
  1998.

\bibitem[{LeCun}(1985)]{lecun:1985}
Y.~{LeCun}.
\newblock A learning scheme for asymmetric threshold networks.
\newblock In \emph{Proceedings of Cognitiva 85}, pages 599--604, Paris, France,
  1985.

\bibitem[{LeCun} et~al.(1998){LeCun}, Bottou, Orr, and M\"uller]{lecun:1998}
Y.~{LeCun}, L.~Bottou, G.~B. Orr, and K.-R. M\"uller.
\newblock Efficient backprop.
\newblock In G.B. Orr and K.-R. M\"uller, editors, \emph{Neural Networks:
  Tricks of the Trade}, pages 9--50. Springer, 1998.

\bibitem[Lewis et~al.(2004)Lewis, Yang, Rose, and Li]{lewis:2004}
D.~D. Lewis, Y.~Yang, T.~G. Rose, and F.~Li.
\newblock Rcv1: A new benchmark collection for text categorization research.
\newblock \emph{Journal of Machine Learning Research}, 5:\penalty0 361--397,
  2004.

\bibitem[Liang(2005)]{liang:2005}
P.~Liang.
\newblock Semi-supervised learning for natural language.
\newblock Master's thesis, Massachusetts Institute of Technology, 2005.

\bibitem[Liang et~al.(2008)Liang, Daum\'{e}, and Klein]{liang:2008}
P.~Liang, H.~Daum\'{e}, III, and D.~Klein.
\newblock Structure compilation: trading structure for features.
\newblock In \emph{International conference on Machine learning (ICML)}, pages
  592--599. ACM, 2008.

\bibitem[Lin and Wu(2009)]{lin:2009}
D.~Lin and X.~Wu.
\newblock Phrase clustering for discriminative learning.
\newblock In \emph{Proceedings of the Association for Computational Linguistics
  (ACL)}, pages 1030--1038. Association for Computational Linguistics, 2009.

\bibitem[Littlestone(1988)]{littlestone:1998}
N.~Littlestone.
\newblock Learning quickly when irrelevant attributes abound: A new
  linear-threshold algorithm.
\newblock In \emph{Machine Learning}, pages 285--318, 1988.

\bibitem[{McCallum} and Li(2003)]{mccallum:2003}
A.~{McCallum} and Wei Li.
\newblock Early results for named entity recognition with conditional random
  fields, feature induction and web-enhanced lexicons.
\newblock In \emph{Proceedings of the seventh conference on Natural language
  learning at HLT-NAACL 2003}, pages 188--191. Association for Computational
  Linguistics, 2003.

\bibitem[McClosky et~al.(2006)McClosky, Charniak, and Johnson]{mcclosky:2008}
D.~McClosky, E.~Charniak, and M.~Johnson.
\newblock {Effective self-training for parsing}.
\newblock \emph{Proceedings of HLT-NAACL 2006}, 2006.

\bibitem[McDonald et~al.(2005)McDonald, Crammer, and Pereira]{mcdonald:2005}
R.~McDonald, K.~Crammer, and F.~Pereira.
\newblock Flexible text segmentation with structured multilabel classification.
\newblock In \emph{HLT '05: Proceedings of the conference on Human Language
  Technology and Empirical Methods in Natural Language Processing}, pages
  987--994. Association for Computational Linguistics, 2005.

\bibitem[Miller et~al.(2000)Miller, Fox, Ramshaw, and Weischedel]{miller:2000}
S.~Miller, H.~Fox, L.~Ramshaw, and R.~Weischedel.
\newblock {A novel use of statistical parsing to extract information from
  text}.
\newblock \emph{6th Applied Natural Language Processing Conference}, 2000.

\bibitem[Miller et~al.(2004)Miller, Guinness, and Zamanian]{miller:2004}
S.~Miller, J.~Guinness, and A.~Zamanian.
\newblock Name tagging with word clusters and discriminative training.
\newblock In \emph{Proceedings of HLT-NAACL}, pages 337--342, 2004.

\bibitem[Mnih and Hinton(2007)]{mnih:2007}
A~Mnih and G.~E. Hinton.
\newblock Three new graphical models for statistical language modelling.
\newblock In \emph{International Conference on Machine Learning, {ICML}}, pages
  641--648, 2007.

\bibitem[Musillo and Merlo(2006)]{musillo:2006}
G.~Musillo and P.~Merlo.
\newblock {Robust Parsing of the Proposition Bank}.
\newblock \emph{ROMAND 2006: Robust Methods in Analysis of Natural language
  Data}, 2006.

\bibitem[Neal(1996)]{neal-1996}
R.~M. Neal.
\newblock \emph{Bayesian Learning for Neural Networks}.
\newblock Number 118 in Lecture Notes in Statistics. Springer-Verlag, New York,
  1996.

\bibitem[Okanohara and Tsujii(2007)]{okanohara:2007}
D.~Okanohara and J.~Tsujii.
\newblock A discriminative language model with pseudo-negative samples.
\newblock \emph{Proceedings of the 45th Annual Meeting of the ACL}, pages
  73--80, 2007.

\bibitem[Palmer et~al.(2005)Palmer, Gildea, and Kingsbury]{propbank}
M.~Palmer, D.~Gildea, and P.~Kingsbury.
\newblock The proposition bank: An annotated corpus of semantic roles.
\newblock \emph{Comput. Linguist.}, 31\penalty0 (1):\penalty0 71--106, 2005.
\newblock ISSN 0891-2017.

\bibitem[Pearl(1988)]{pearl:1988}
J.~Pearl.
\newblock \emph{Probabilistic Reasoning in Intelligent Systems}.
\newblock Morgan Kaufman, San Mateo, 1988.

\bibitem[Plaut and Hinton(1987)]{plaut:1987}
D.~C. Plaut and G.~E. Hinton.
\newblock Learning sets of filters using back-propagation.
\newblock \emph{Computer Speech and Language}, 2:\penalty0 35--61, 1987.

\bibitem[Porter(1980)]{porter:1980}
M.~F. Porter.
\newblock An algorithm for suffix stripping.
\newblock \emph{Program}, 14\penalty0 (3):\penalty0 130--137, 1980.

\bibitem[Pradhan et~al.(2004)Pradhan, Ward, Hacioglu, Martin, and
  Jurafsky]{pradhan2004ssp}
S.~Pradhan, W.~Ward, K.~Hacioglu, J.~Martin, and D.~Jurafsky.
\newblock {Shallow semantic parsing using support vector machines}.
\newblock \emph{Proceedings of HLT/NAACL-2004}, 2004.

\bibitem[Pradhan et~al.(2005)Pradhan, Hacioglu, Ward, Martin, and
  Jurafsky]{pradhan:2005}
S.~Pradhan, K.~Hacioglu, W.~Ward, J.~H. Martin, and D.~Jurafsky.
\newblock Semantic role chunking combining complementary syntactic views.
\newblock In \emph{Proceedings of the Ninth Conference on Computational Natural
  Language Learning (CoNLL-2005)}, pages 217--220. Association for
  Computational Linguistics, June 2005.

\bibitem[Punyakanok et~al.(2005)Punyakanok, Roth, and Yih]{punyakanok:2005}
V.~Punyakanok, D.~Roth, and W.~Yih.
\newblock The necessity of syntactic parsing for semantic role labeling.
\newblock In \emph{IJCAI}, pages 1117--1123, 2005.

\bibitem[Ratinov and Roth(2009)]{ratinov:2009}
L.~Ratinov and D.~Roth.
\newblock Design challenges and misconceptions in named entity recognition.
\newblock In \emph{Proceedings of the Thirteenth Conference on Computational
  Natural Language Learning (CoNLL)}, pages 147--155. Association for
  Computational Linguistics, 2009.

\bibitem[Ratnaparkhi(1996)]{ratnaparkhi:1996}
A.~Ratnaparkhi.
\newblock A maximum entropy model for part-of-speech tagging.
\newblock In Eric Brill and Kenneth Church, editors, \emph{Proceedings of the
  Conference on Empirical Methods in Natural Language Processing}, pages
  133--142. Association for Computational Linguistics, 1996.

\bibitem[Rosenfeld and Feldman(2007)]{rosenfeld:2007}
B.~Rosenfeld and R.~Feldman.
\newblock {Using Corpus Statistics on Entities to Improve Semi-supervised
  Relation Extraction from the Web}.
\newblock \emph{Proceedings of the 45th Annual Meeting of the ACL}, pages
  600--607, 2007.

\bibitem[Rumelhart et~al.(1986)Rumelhart, Hinton, and Williams]{rumelhart:1986}
D.~E. Rumelhart, G.~E. Hinton, and R.~J. Williams.
\newblock Learning internal representations by back-propagating errors.
\newblock In D.E. Rumelhart and J.~L. McClelland, editors, \emph{Parallel
  Distributed Processing: Explorations in the Microstructure of Cognition},
  volume~1, pages 318--362. {MIT} Press, 1986.

\bibitem[Sch\"{u}tze(1995)]{schutze:1995}
H.~Sch\"{u}tze.
\newblock Distributional part-of-speech tagging.
\newblock In \emph{Proceedings of the Association for Computational Linguistics
  (ACL)}, pages 141--148. Morgan Kaufmann Publishers Inc., 1995.

\bibitem[Schwenk and Gauvain(2002)]{schwenk:2002}
H.~Schwenk and J.~L. Gauvain.
\newblock Connectionist language modeling for large vocabulary continuous
  speech recognition.
\newblock In \emph{IEEE International Conference on Acoustics, Speech, and
  Signal Processing}, pages 765--768, 2002.

\bibitem[Sha and Pereira(2003)]{sha:2003}
F.~Sha and F.~Pereira.
\newblock Shallow parsing with conditional random fields.
\newblock In \emph{NAACL '03: Proceedings of the 2003 Conference of the North
  American Chapter of the Association for Computational Linguistics on Human
  Language Technology}, pages 134--141. Association for Computational
  Linguistics, 2003.

\bibitem[Shannon(1951)]{shannon:1951}
C.~E. Shannon.
\newblock Prediction and entropy of printed english.
\newblock \emph{Bell Systems Technical Journal}, 30:\penalty0 50--64, 1951.

\bibitem[Shen and Sarkar(2005)]{shen:2005}
H.~Shen and A.~Sarkar.
\newblock Voting between multiple data representations for text chunking.
\newblock \emph{Advances in Artificial Intelligence}, pages 389--400, 2005.

\bibitem[Shen et~al.(2007)Shen, Satta, and Joshi]{shen:2007}
L.~Shen, G.~Satta, and A.~K. Joshi.
\newblock Guided learning for bidirectional sequence classification.
\newblock In \emph{Proceedings of the 45th Annual Meeting of the Association
  for Computational Linguistics (ACL)}, 2007.

\bibitem[Smith and Eisner(2005)]{smith:2005}
N.~A. Smith and J.~Eisner.
\newblock Contrastive estimation: Training log-linear models on unlabeled data.
\newblock In \emph{Proceedings of the 43rd Annual Meeting of the Association
  for Computational Linguistics (ACL)}, pages 354--362. Association for
  Computational Linguistics, 2005.

\bibitem[Suddarth and Holden(1991)]{suddarth-1991}
S.~C. Suddarth and A.~D.~C. Holden.
\newblock Symbolic-neural systems and the use of hints for developing complex
  systems.
\newblock \emph{International Journal of Man-Machine Studies}, 35\penalty0
  (3):\penalty0 291--311, 1991.

\bibitem[Sun et~al.(2008)Sun, Morency, Okanohara, and Tsujii]{sun:2008}
X.~Sun, L.-P. Morency, D.~Okanohara, and J.~Tsujii.
\newblock Modeling latent-dynamic in shallow parsing: a latent conditional
  model with improved inference.
\newblock In \emph{COLING '08: Proceedings of the 22nd International Conference
  on Computational Linguistics}, pages 841--848. Association for Computational
  Linguistics, 2008.

\bibitem[Sutton and McCallum(2005{\natexlab{a}})]{sutton:2005}
C.~Sutton and A.~McCallum.
\newblock Joint parsing and semantic role labeling.
\newblock In \emph{Proceedings of CoNLL-2005}, pages 225--228,
  2005{\natexlab{a}}.

\bibitem[Sutton and McCallum(2005{\natexlab{b}})]{sutton:2005a}
C.~Sutton and A.~McCallum.
\newblock {Composition of conditional random fields for transfer learning}.
\newblock \emph{Proceedings of the conference on Human Language Technology and
  Empirical Methods in Natural Language Processing}, pages 748--754,
  2005{\natexlab{b}}.

\bibitem[Sutton et~al.(2007)Sutton, McCallum, and Rohanimanesh]{sutton:2007}
C.~Sutton, A.~McCallum, and K.~Rohanimanesh.
\newblock {Dynamic Conditional Random Fields: Factorized Probabilistic Models
  for Labeling and Segmenting Sequence Data}.
\newblock \emph{JMLR}, 8:\penalty0 693--723, 2007.

\bibitem[Suzuki and Isozaki(2008)]{suzuki:2008}
J.~Suzuki and H.~Isozaki.
\newblock Semi-supervised sequential labeling and segmentation using giga-word
  scale unlabeled data.
\newblock In \emph{Proceedings of ACL-08: HLT}, pages 665--673, Columbus, Ohio,
  June 2008. Association for Computational Linguistics.

\bibitem[Teahan and Cleary(1996)]{teahan-1996}
W.~J. Teahan and J.~G. Cleary.
\newblock The entropy of english using ppm-based models.
\newblock In \emph{In Data Compression Conference ({DCC'96})}, pages 53--62.
  IEEE Computer Society Press, 1996.

\bibitem[Toutanova et~al.(2003)Toutanova, Klein, Manning, and
  Singer]{toutanova:2003}
K.~Toutanova, D.~Klein, C.~D. Manning, and Y.~Singer.
\newblock Feature-rich part-of-speech tagging with a cyclic dependency network.
\newblock In \emph{HLT-NAACL}, 2003.

\bibitem[Turian et~al.(2010)Turian, Ratinov, and Bengio]{turian:2010}
J.~Turian, L.~Ratinov, and Y.~Bengio.
\newblock Word representations: A simple and general method for semi-supervised
  learning.
\newblock In \emph{Proceedings of the Association for Computational Linguistics
  (ACL)}, pages 384--392. Association for Computational Linguistics, 2010.

\bibitem[Ueffing et~al.(2007)Ueffing, Haffari, and Sarkar]{ueffing:2007}
N.~Ueffing, G.~Haffari, and A.~Sarkar.
\newblock {Transductive learning for statistical machine translation}.
\newblock \emph{Proceedings of the 45th Annual Meeting of the ACL}, pages
  25--32, 2007.

\bibitem[Waibel et~al.(1989)Waibel, Hanazawa, Hinton, Shikano, and
  Lang]{waibel:1989}
A.~Waibel, T.~Hanazawa, G.~Hinton, K.~Shikano, and K.J. Lang.
\newblock Phoneme recognition using time-delay neural networks.
\newblock \emph{{IEEE} Transactions on Acoustics, Speech, and Signal
  Processing}, 37\penalty0 (3):\penalty0 328--339, 1989.

\bibitem[Weston et~al.(2008)Weston, Ratle, and Collobert]{weston2008deep}
J.~Weston, F.~Ratle, and R.~Collobert.
\newblock {Deep learning via semi-supervised embedding}.
\newblock In \emph{Proceedings of the 25th international conference on Machine
  learning}, pages 1168--1175. ACM, 2008.

\end{thebibliography}

\end{document}